\documentclass[lettersize,journal]{IEEEtran}
\usepackage{amsmath,amsfonts}
\usepackage{algorithmic}
\usepackage{array}
\usepackage{cite}
\usepackage[caption=false,font=normalsize,labelfont=sf,textfont=sf]{subfig}
\usepackage{textcomp}
\usepackage{stfloats}
\usepackage{url}
\usepackage{verbatim}
\usepackage{booktabs}
\usepackage{multirow}
\usepackage{graphicx}
\usepackage{makecell}
\def\BibTeX{{\rm B\kern-.05em{\sc i\kern-.025em b}\kern-.08em
    T\kern-.1667em\lower.7ex\hbox{E}\kern-.125emX}}
\usepackage{balance}
\begin{document}

\title{MADSNet: A Multi-Scale Axial-Attentive Dense Spiking Network for Spike Camera Image Reconstruction}

\author{
Sirui Li,
Rui Zhao, \textit{Member, IEEE},
Siqi Cai, \textit{Senior Member, IEEE},
Haizhou Li, \textit{Fellow, IEEE}
\\
Tiejun Huang, \textit{Senior Member, IEEE},
Yajing Zheng, \textit{Member, IEEE}
\thanks{
S. Li is with the  School of Data Science, The Chinese University of Hong Kong, Shenzhen, China, and this work was conducted during his internship at the School of Computer Science, Peking University, Beijing, China. E-mail: 223040027@link.cuhk.edu.cn.
}
\thanks{
R. Zhao is with the College of Computing and Data Science, Nanyang Technological University, Singapore. E-mail: ruizhao26@gmail.com.
}
\thanks{
S. Cai is with the School of Intelligence Science and Engineering, Harbin Institute of Technology, Shenzhen, China. E-mail: caisiqi@ieee.org.
}
\thanks{
H. Li is with the  School of Data Science, The Chinese University of Hong Kong, Shenzhen, China. E-mail: haizhouli@cuhk.edu.cn.
}
\thanks{
T. Huang and Y. Zheng are with the State Key Laboratory for Multimedia Information Processing, School of Computer Science, Peking University, Beijing, China. 
E-mail: tjhuang@pku.edu.cn, yj.zheng@pku.edu.cn.
}
}
\maketitle

\markboth{IEEE TRANSACTIONS ON IMAGE PROCESSING}%
{Shell \MakeLowercase{\textit{et al.}}: Bare Demo of IEEEtran.cls for Computer Society Journals}

\begin{abstract}
Spike cameras encode dynamic scenes into high-temporal-resolution binary spike streams for efficient high-speed vision. Although spiking neural networks (SNNs) can process such signals with sparse and accumulate-only computation, existing SNN-based reconstruction methods are typically shallow and still lag behind ANN-based models in reconstruction fidelity.
This paper proposes MADSNet, a Multi-scale Axial-attentive Dense Spiking Network for spike camera image reconstruction. MADSNet adopts a hierarchical spiking encoder--decoder, where a multi-scale spike stem extracts early features, axial spike attention captures long-range dependencies, a dense spiking bottleneck enhances feature reuse, and a spike-to-continuous decoder reconstructs continuous intensity images from spike-domain representations.
Experiments on gray-scale and color spike reconstruction show that MADSNet outperforms SSIR by 4.3 dB PSNR and Spk2ImgNet by 1.94 dB on SREDS. Compared with Spk2ImgNet, it achieves about 15$\times$ lower latency and uses only about 1/40 of the estimated energy, demonstrating that carefully designed SNN architectures can improve reconstruction quality while preserving computational efficiency.
\end{abstract}


\section{Introduction}

\IEEEPARstart{C}{onventional} frame-based cameras capture visual scenes by integrating photons
within a fixed exposure interval. This imaging mechanism works well for
ordinary scenes, but it inevitably suffers from motion blur, temporal
discretization, and missing inter-frame information when recording high-speed
dynamic processes. Spike cameras provide a neuromorphic alternative by
representing visual signals as continuous binary spike streams
~\cite{huang20231000,zheng2026spikecv}. In a spike camera, each
pixel independently accumulates incoming light intensity and asynchronously
fires a binary spike once the accumulated signal reaches a predefined
threshold, after which the accumulator is reset for the next integration cycle.
This integrate-and-fire sensing mechanism enables ultra-high temporal
resolution, e.g., up to tens of thousands of spike frames per second, while
preserving dense temporal information for dynamic scene analysis
~\cite{zhu2019retina,huang20231000,zheng2026spikecv}.

The development of spike cameras has stimulated a broad range of visual
applications, including high-speed image reconstruction~\cite{zhao2021reconstructing,zhao2021spk2imgnet,zhao2023spike,
chen2022self, zhao2024boosting}, motion deblurring~\cite{chen2023self,zhang2024spike}, optical flow estimation~\cite{hu2022optical,zhao2022learning,zhao2024optical,zhao2025spike}, tracking~\cite{zhu2022fpga,zheng2026snntracker}, joint visual compression and perception~\cite{feng2025joint},
and color spike imaging~\cite{dong20223d,dong2025self,dong2024learning,dong2025color}. In parallel, the SpikeCV platform has been developed to
support the spike vision ecosystem by providing spike datasets, hardware
interfaces, data loaders, algorithm libraries, and unified processing pipelines
for different spike-based visual tasks~\cite{zheng2026spikecv}. These efforts
indicate that spike cameras are evolving from a specialized high-speed imaging
sensor into a broader computational vision platform.

Within this ecosystem, spike-to-image reconstruction occupies a foundational position: it converts raw spike streams into the intensity representations that human visualization and most downstream vision tasks still depend on. Yet recovering high-quality intensity remains challenging, as the task is inherently ill-posed: the network must infer continuous scene radiance from observations that are sparse, binary, temporally accumulated, and corrupted by photon-arrival noise and motion. Early reconstruction methods, such as TFP~\cite{zhu2019retina}, TFI~\cite{zhu2019retina}, TFSTP~\cite{zheng2021high}, and MAHTF~\cite{zhao2021reconstructing}, estimate image intensity from hand-crafted spike statistics or temporal filtering. Although these methods are computationally simple, they are limited in recovering high-fidelity textures and complex spatiotemporal structures. Learning-based methods, including Spk2ImgNet and recent self-supervised reconstruction frameworks, significantly improve reconstruction quality by exploiting data-driven feature learning and motion-aware supervision~\cite{zhao2021spk2imgnet,chen2022self,chen2023self,zhao2024boosting,chen2025spikeclip}. However, most of these methods are based on artificial neural networks (ANNs), whose dense multiply-accumulate operations lead to high latency and energy consumption.

Spiking neural networks (SNNs)~\cite{maass1997networks} provide a natural computational counterpart to spike cameras. By processing spike-domain activations with event-driven accumulate-only operations, SNNs have the potential to realize low-power and low-latency for spike cameras~\cite{zheng2022spike,zheng2023capture,zhao2023spike,zhao2024recognizing}. SSIR~\cite{zhao2023spike} is a representative SNN-based reconstruction method that combines Spike-Element-Wise (SEW) residual blocks with a reparameterizable Progressive Filtering Module (PFM), reporting substantially lower energy consumption than ANN-based reconstruction. However, there remains a clear reconstruction-quality gap between existing SNN-based methods and high-capacity ANN models. We argue that this gap is not solely caused by the spiking neuron model, but also by insufficient architecture design. In particular, existing SNN reconstruction networks typically use single-scale early embeddings that overlook multi-scale spatial structure, rely on spike-domain encoders confined to local convolutions that cannot capture long-range dependencies, and adopt relatively direct decoding heads that handle the transition from discrete spike activations to continuous intensities only crudely.

To address these limitations, we propose MADSNet, a multi-scale axial-attentive dense spiking network for spike camera image reconstruction. The core design principle of MADSNet is to improve representation capacity while preserving the accumulate-only, multiplication-free computation that underlies the energy efficiency of SNNs. Specifically, MADSNet introduces a Multi-Scale Spike Stem (MS3Stem) to jointly capture inter-channel temporal correlations and local-to-large spatial patterns from stacked spike frames. It further employs depthwise-separable SEW blocks and light spike blocks with axial spike attention to build an efficient spike-domain encoder. A dense spike bottleneck is introduced to enhance latent feature reuse and stabilize information propagation in deeper spiking architectures. Finally, spike pixel-shuffle upsampling and a spike-to-continuous fusion block are used to recover fine spatial details and convert discrete spike-domain features into continuous intensity predictions.


MADSNet achieves a favorable balance between reconstruction accuracy and computational efficiency. On the SREDS benchmark, MADSNet reaches 37.232 dB PSNR, improving over the SNN baseline SSIR by +4.30 dB and surpassing the ANN-based Spk2ImgNet by +1.94 dB. At the same time, MADSNet remains much more efficient than Spk2ImgNet, achieving about 15$\times$ lower inference latency, 3.4$\times$ lower GPU memory usage, and only about 1/40 of the estimated energy consumption. These results suggest that SNN-based reconstruction does not have to trade substantial accuracy for efficiency; instead, with appropriate architecture design, SNNs can provide a strong accuracy--efficiency balance for spike camera reconstruction.



The main contributions of this work are threefold:
\begin{itemize}
    \item We propose MADSNet, a hierarchical spiking encoder-decoder built on the design principle of improving representation capacity while retaining the accumulate-only efficiency of spike-domain computation.
    \item We develop a coherent set of spike-oriented modules that progressively enrich the spike-domain representation along the entire pipeline, from multi-scale early embedding, through axial-attentive encoding and dense latent aggregation, to spike-to-continuous decoding.
    \item We conduct experiments on gray-scale and color spike reconstruction, demonstrating superior accuracy and efficiency trade-off over existing ANN and SNN baselines.
\end{itemize}

\section{Related Work}

\subsection{Spike and Event Camera Image Reconstruction}

\subsubsection{Gray-Scale Spike Camera Reconstruction}
Early filtering-based methods estimate intensity directly from spike statistics without learning. TFP and TFI~\cite{zhu2019retina} infer pixel intensity from spike firing frequency and inter-spike intervals, respectively, while TFSTP~\cite{zheng2021high} and MAHTF~\cite{zhao2021reconstructing,zheng2023capture} exploit short-term plasticity and motion-aligned temporal filtering for higher fidelity, albeit at high latency. Learning-based methods improve quality further: Spk2ImgNet~\cite{zhao2021spk2imgnet} aligns multi-temporal spike features with deformable convolutions but is computationally heavy, whereas SSIR~\cite{zhao2023spike} proposes the first SNN-based reconstruction with SEW residual blocks and a reparameterizable PFM, reducing energy to about $1/1019$ of Spk2ImgNet but still trailing high-capacity ANNs in fidelity.

\subsubsection{Color Spike Camera Reconstruction}
The color spike camera (CSC) employs a Bayer-pattern Color Filter Array, introducing the additional challenges of spatially missing pixels per channel and inter-channel chromatic correlations. 3DRI~\cite{dong20223d} fills missing pixels via 3D residual interpolation using spatially and temporally neighboring samples, while subsequent work has developed deep demosaicing networks explicitly tailored for spike camera CFAs~\cite{dong2024learning}. SJRE~\cite{chen2023self}, a self-supervised gray-scale method, can be adapted to color streams by processing each Bayer channel independently, but lacks explicit inter-channel modeling. SSL-CSC~\cite{dong2025self} proposes a motion-guided pipeline that jointly estimates optical flow across color channels and trains a self-supervised network with a masked spike encoder to suppress quantization noise, establishing the current state-of-the-art among methods that do not require ground-truth supervision. Additionally, recent advances have demonstrated that aggregating long short-term temporal spike signals can further enhance continuous color reconstruction fidelity~\cite{dong2025color}. These methods, however, are purpose-built for color reconstruction rather than transferred from a general spike architecture.

\subsubsection{Event Camera Reconstruction}
A closely related neuromorphic modality is the event camera, which records per-pixel asynchronous brightness changes as a stream of events, enabling high dynamic range and microsecond temporal resolution~\cite{gallego2020event}. Because event and spike cameras both represent visual information as asynchronous data streams, their deep-learning reconstruction pipelines share methodological similarities, and event-to-video reconstruction has demonstrated significant progress. E2VID~\cite{rebecq2019events} is a seminal work for this purpose, employing a recurrent fully convolutional network to convert event streams into video frames. Following E2VID, numerous studies have augmented this paradigm from various angles, including FireNet~\cite{scheerlinck2020fast}, E2VID++~\cite{stoffregen2020reducing}, and FireNet++~\cite{stoffregen2020reducing}. ETNet~\cite{weng2021event} further incorporates Transformer-based attention into event-to-video reconstruction, improving the modeling of long-range temporal dependencies. HyperE2VID~\cite{ercan2024hypere2vid} adopts a dynamic neural network architecture and uses hypernetworks to generate per-pixel adaptive filters, achieving state-of-the-art reconstruction quality. However, event cameras record only relative intensity changes and thus cannot recover absolute texture detail, making event-based reconstruction complementary to but distinct from the spike-based reconstruction discussed above.

\subsection{Spiking Neurons in SNNs}

The Integrate-and-Fire (IF) neuron~\cite{maass1997networks} fires a binary spike
when its accumulated membrane potential exceeds a threshold, offering
event-driven efficiency but limited temporal expressiveness.
The Leaky Integrate-and-Fire (LIF) neuron~\cite{wu2018spatio} introduces
membrane potential decay, enabling implicit short-term memory.
To further improve adaptability, Parametric LIF (PLIF)~\cite{fang2021incorporating}
makes the time constant learnable, which is adopted in SSIR
to improve training convergence.
However, direct training via spatio-temporal backpropagation requires
storing neuronal states across all timesteps, imposing significant GPU
memory overhead, while the non-differentiable spike generation function
necessitates surrogate gradient approximations that introduce
inherent estimation errors and hinder stable convergence~\cite{li2021differentiable}.
To address this, I-LIF~\cite{luo2024integer} allows integer-valued spike
emission during training while decomposing to binary spikes at inference,
preserving accumulate-only (AC) computation on neuromorphic hardware.
IBRA-LIF~\cite{ye2025integer} further introduces a Range Alignment strategy
that scales the effective dynamic range of integer spikes, alleviating
quantization errors that accumulate across deep layers.

These neuron-level developments improve the representational capacity and
training stability of SNNs, but they are orthogonal to architecture-level
designs for spike reconstruction. In this work, we adopt IBRA-LIF as the basic
activation function for spiking layers and focus on designing an efficient
encoder-decoder architecture for gray-scale spike reconstruction.

\section{Method}
\subsection{Spike Generation Mechanism}

A spike camera sensor consists of an array of pixels that each independently
accumulate incident photons in a continuous, asynchronous manner, mimicking
the integrate-and-fire (I\&F) dynamics of biological retinal ganglion cells.
Specifically, each pixel maintains an internal accumulator whose state at
spatial location $\mathbf{x}$ and time $t$ evolves as
\begin{equation}
    A(\mathbf{x},\,t)
    = \int_{0}^{t} \alpha \cdot I(\mathbf{x},\,\tau)\,\mathrm{d}\tau
    \;\;\mathrm{mod}\;\theta,
    \label{eq:accum}
\end{equation}
where $I(\mathbf{x},t)$ is the scene radiance (light intensity), $\alpha$ is
the quantum efficiency of the photodetector, and $\theta$ is the fixed firing
threshold. Whenever the accumulated value reaches $\theta$, the pixel fires a
binary spike and immediately resets its accumulator to zero, starting the next
integration cycle.

The sensor output is read out at a fixed interval $\Delta t$.
Denoting the $k$-th readout window as
$\mathcal{T}_k := \bigl((k{-}1)\Delta t,\, k\Delta t\bigr]$,
the $k$-th binary spike frame records whether at least one firing event
occurred within that window:
\begin{equation}
    S_{k}(\mathbf{x})
    = \mathbf{1}\!\left[
        \exists\; t \in \mathcal{T}_{k} :\;
        A(\mathbf{x},\,t) = \theta
    \right],
    \label{eq:fire}
\end{equation}
where $k = 1,\ldots,T$, and the accumulator is immediately reset,
$A(\mathbf{x},\,t)\leftarrow 0$, upon each firing event.
Because the inter-spike interval at each pixel is inversely proportional to
the local light intensity, brighter regions produce denser spike trains whereas
darker regions produce sparser ones. This design enables ultra-high temporal
resolution (up to tens of thousands of spike frames per second) and avoids the
motion blur and temporal aliasing of conventional frame-based exposure.
In practice, the continuous spike stream is discretized into $T$ binary frames
$\mathbf{S}\in\{0,1\}^{T\times H\times W}$ over a short temporal window of
duration $T\Delta t$, which serves as the input to the reconstruction network.
Recovering the continuous intensity image $\hat{\mathbf{I}}\in[0,1]^{H\times W}$
from this sparse binary observation is inherently ill-posed, as a single spike
sequence is consistent with a range of underlying radiance fields; this
motivates the network design described in the following subsections.
\subsection{Overall Architecture}

Given a sequence of $T$ binary spike frames
$\mathbf{S}\in\{0,1\}^{T\times H\times W}$, the goal of gray-scale spike
reconstruction is to recover the corresponding intensity image
$\hat{\mathbf{I}}\in[0,1]^{H\times W}$. Unlike frame-based image restoration,
spike reconstruction requires the network to infer continuous intensity values
from sparse, asynchronous, and temporally accumulated binary observations.
This poses three main challenges: effective early spike representation,
efficient spike-domain feature learning, and smooth conversion from discrete
spiking activations to continuous image intensities.

\begin{figure*}[t]
    \centering
    \includegraphics[width=\textwidth]{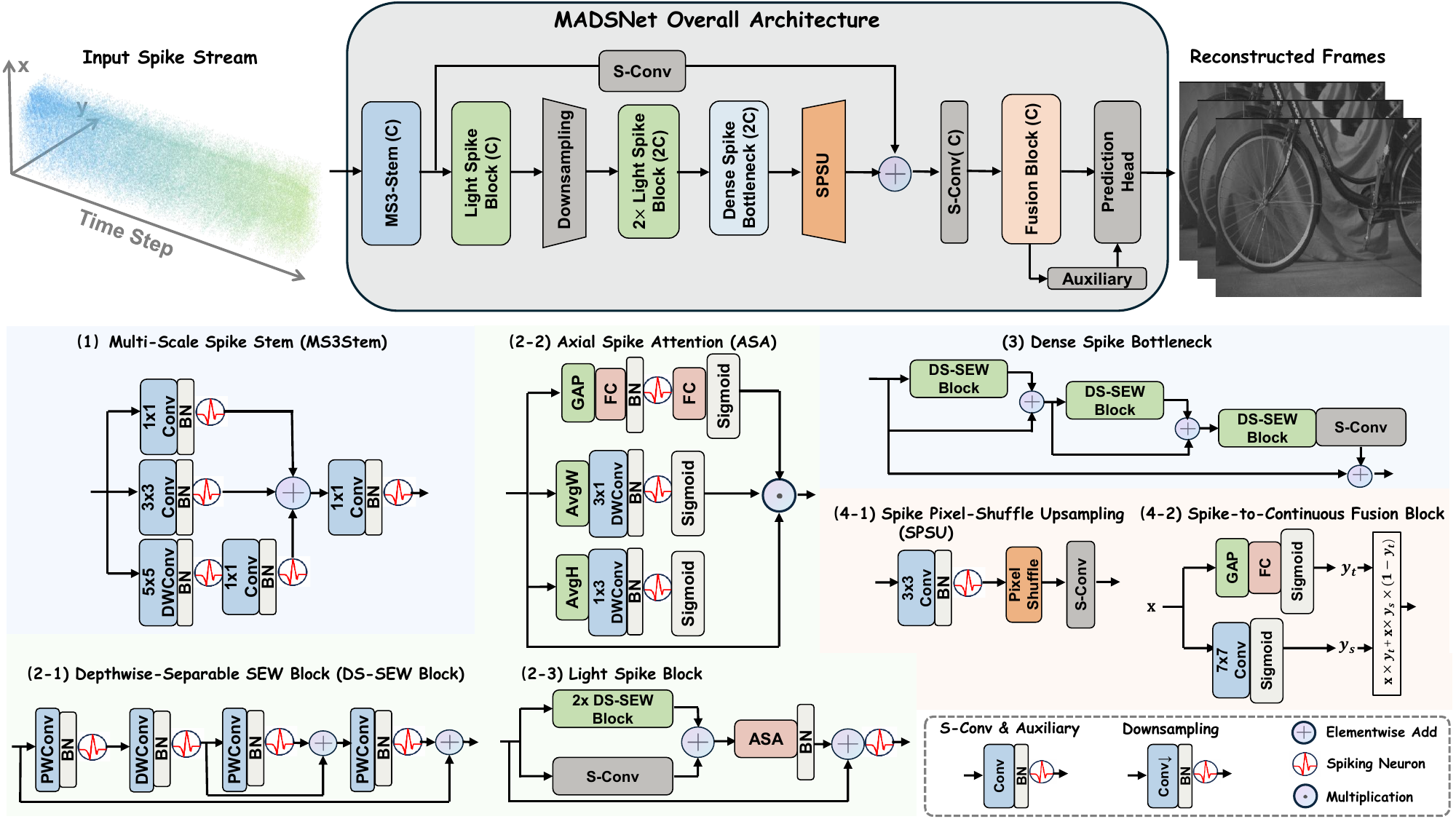}
    \caption{Overall architecture of MADSNet and detailed diagrams of each
    sub-module. \textbf{Top:} the full encoder-decoder pipeline from input
    spike stream to reconstructed frames. \textbf{Bottom:} detailed structures
    of MS3Stem,DS-SEW Block,Axial Spike
    Attention (ASA), Light Spike Block, Dense Spike Bottleneck,   Spike Pixel-Shuffle Upsampling (SPSU),
    and the Spike-to-Continuous Fusion Block.}
    \label{fig:architecture}
\end{figure*}

To address these challenges, we propose MADSNet, a hierarchical spiking
encoder-decoder architecture. As illustrated in Fig.~\ref{fig:architecture},
MADSNet first employs a Multi-Scale Spike Stem (MS3Stem) to embed the input
spike sequence into a rich spike-domain representation. The encoder then uses
light spike blocks to jointly model local structures and long-range spatial
dependencies with low computational cost. A dense spike bottleneck is inserted
at the latent stage to enhance feature reuse and stabilize information
propagation. In the decoder, Spike Pixel-Shuffle Upsampling (SPSU) restores
spatial resolution, while a Spike-to-Continuous Fusion Block bridges the
discrete spiking backbone and the continuous prediction head. An auxiliary
prediction branch is used only during training to provide deep supervision.

All spiking convolutional layers in MADSNet are built upon the IBRA-LIF neuron,
which improves spike representation during training while preserving
accumulate-only inference through binary decomposition.

Unless otherwise specified, all spiking convolutional layers use the IBRA-LIF
neuron~\cite{ye2025integer} as the activation function. IBRA-LIF provides
range-aligned integer spike representations during training and supports
binary spike decomposition during inference, preserving the accumulate-only
computation property of SNNs. Since the neuron model is not the main
architectural contribution of this work, we follow the original setting
of~\cite{ye2025integer} and focus the following description on the proposed
architecture.

\subsection{Multi-Scale Spike Stem}

The original SSIR starts from a single spiking convolution that captures
mainly local spatial patterns.
However, the stacked spike tensor encodes temporal cues in its channel
dimension while spatial patterns jointly determine the target intensity image.
To exploit this property, we design a three-branch multi-scale spike stem.
 
Specifically, the input spike tensor $\mathbf{X}$ is processed in parallel by:
\begin{align}
\mathbf{f}_{\text{pt}}  &= \phi\!\left(\text{BN}(\text{Conv}_{1\times1}(\mathbf{X}))\right),\\
\mathbf{f}_{\text{loc}} &= \phi\!\left(\text{BN}(\text{Conv}_{3\times3}(\mathbf{X}))\right),\\
\mathbf{f}_{\text{lk}}  &= \phi\!\left(\text{BN}\!\left(\text{Conv}_{1\times1}\!\left(
                           \phi(\text{BN}(\text{DWConv}_{5\times5}(\mathbf{X})))\right)\right)\right),
\end{align}
where $\phi(\cdot)$ denotes the IBRA-LIF neuron.
The three branches are concatenated and fused by a $1\!\times\!1$ spiking convolution:
\begin{equation}
\mathbf{s}_{0} = \phi\!\left(\text{BN}\!\left(\text{Conv}_{1\times1}
([\mathbf{f}_{\text{pt}},\,\mathbf{f}_{\text{loc}},\,\mathbf{f}_{\text{lk}}])\right)\right).
\end{equation}
 
The $1\!\times\!1$ branch models inter-channel (temporal) correlations,
the $3\!\times\!3$ branch captures local spatial textures,
and the depthwise-separable $5\!\times\!5$ branch enlarges the receptive field
with limited overhead.
This design makes the first-stage spike embedding significantly richer than the
single-branch baseline of SSIR.
 
\subsection{Efficient Spike-Domain Encoding}

\subsubsection{Depthwise-Separable SEW Block}
 
The original SSIR adopts a standard bottleneck SEW block.
To reduce computation while maintaining residual learning capability,
we redesign it as a depthwise-separable SEW block.
Given an input feature $\mathbf{x}$ with $C$ channels, we first compress
the channel dimension:
\begin{equation}
\mathbf{u}_{1} = \phi\!\left(\text{BN}(\text{PW}_{1}(\mathbf{x}))\right),
\end{equation}
where $\text{PW}_{1}$ is a $1\!\times\!1$ pointwise convolution that projects
to $C/4$ channels.
Spatial interaction is then modelled by a depthwise convolution:
\begin{equation}
\mathbf{u}_{2} = \phi\!\left(\text{BN}(\text{DW}_{3\times3}(\mathbf{u}_{1}))\right).
\end{equation}
An inner micro-residual path is introduced to enhance feature propagation:
\begin{equation}
\mathbf{u}_{3} = \phi\!\left(\text{BN}(\text{PW}_{\text{in}}(\mathbf{u}_{2}))\right)
               + \mathbf{u}_{2}.
\end{equation}
Finally, the feature is projected back to $C$ channels and added to the input:
\begin{equation}
\mathbf{y} = \phi\!\left(\text{BN}(\text{PW}_{\text{out}}(\mathbf{u}_{3}))\right)
           + \mathbf{x}.
\end{equation}
 
Compared with the original bottleneck block, the proposed DS-SEWBlock decouples
spatial and channel mixing, introduces an additional internal gradient path,
and substantially reduces parameter count.
 
\subsubsection{Axial Spike Attention(ASA)}

To model long-range spatial dependencies efficiently, we propose an axial spike
attention module comprising one channel attention branch and two orthogonal
axial spatial branches. Instead of densely computing attention over the full
2-D spatial grid, it factorizes spatial modeling into two 1-D axes.

For channel attention, the input feature is squeezed by global average pooling
and processed by two fully-connected layers:
\begin{equation}
\mathbf{m}_{c} = \sigma\!\left(\text{FC}_{2}\!\left(
\phi(\text{BN}(\text{FC}_{1}(\text{GAP}(\mathbf{x}))))\right)\right),
\end{equation}
where $\sigma(\cdot)$ denotes the sigmoid function.

For spatial attention, the feature is aggregated along the horizontal and
vertical axes independently:
\begin{align}
\mathbf{m}_{h} &= \sigma\!\left(
    \phi\!\left(\text{BN}(\text{DWConv}_{3\times1}(\text{Avg}_{W}(\mathbf{x})))\right)\right),\\
\mathbf{m}_{w} &= \sigma\!\left(
    \phi\!\left(\text{BN}(\text{DWConv}_{1\times3}(\text{Avg}_{H}(\mathbf{x})))\right)\right).
\end{align}
The final attention-modulated feature is
\begin{equation}
\mathbf{y} = \mathbf{x} \odot \mathbf{m}_{c} \odot \mathbf{m}_{h} \odot \mathbf{m}_{w},
\end{equation}
where $\odot$ denotes element-wise multiplication.

This factorised design captures spatial context along two orthogonal axes,
generating the spatial attention at $\mathcal{O}((H+W)\cdot C)$ cost rather than
the $\mathcal{O}(H\cdot W\cdot C\cdot k^{2})$ of producing a dense 2-D attention
map, thus substantially reducing overhead, while the two axial branches provide
global horizontal and vertical context that complements the local receptive
field of the depthwise-separable convolutions.

\subsubsection{Light Spike Block}
 
To build an efficient encoder stage, we introduce a light spike block that
combines local residual modelling and attention modulation.
Two parallel branches are computed:
\begin{align}
\mathbf{f}_{1} &= \mathcal{D}_{2}(\mathcal{D}_{1}(\mathbf{x})),\\
\mathbf{f}_{2} &= \mathcal{S}_{3\times3}(\mathbf{x}),
\end{align}
where $\mathcal{D}_{1},\mathcal{D}_{2}$ are DS-SEWBlocks and
$\mathcal{S}_{3\times3}$ is a standard spiking convolution.
The branches are aggregated and refined by ASA:
\begin{equation}
\mathbf{f} = \mathcal{A}(\mathbf{f}_{1} + \mathbf{f}_{2}),
\end{equation}
and the block output is formed as a spike-domain residual:
\begin{equation}
\mathbf{y} = \phi\!\left(\mathbf{x} + \text{BN}(\mathbf{f})\right).
\end{equation}

\subsection{Dense Spike Bottleneck}

To improve feature reuse in the latent space, we introduce a dense spike
bottleneck composed of three DS-SEWBlocks
$\{\mathcal{B}_{1},\mathcal{B}_{2},\mathcal{B}_{3}\}$.
Rather than stacking blocks sequentially, we design a directed dense aggregation
built on additive cross-block skips:
\begin{align}
\mathbf{x}_{1} &= \mathcal{B}_{1}(\mathbf{x}_{0}),\\
\mathbf{x}_{2} &= \mathcal{B}_{2}(\mathbf{x}_{0} + \mathbf{x}_{1}),\\
\mathbf{x}_{3} &= \mathcal{B}_{3}(\mathbf{x}_{1} + \mathbf{x}_{2}).
\end{align}
The bottleneck output is obtained by a residual projection:
\begin{equation}
\mathbf{y} = \mathbf{x}_{0}
           + \phi\!\left(\text{BN}(\text{Conv}_{1\times1}(\mathbf{x}_{3}))\right).
\end{equation}

In contrast to the concatenation-based connectivity of DenseNet, these skips are
additive and thus introduce no channel growth, keeping the bottleneck
lightweight. The design lets $\mathcal{B}_{2}$ re-access the initial feature
$\mathbf{x}_{0}$ to prevent information forgetting, while $\mathcal{B}_{3}$
aggregates the newly produced representations $\mathbf{x}_{1}$ and
$\mathbf{x}_{2}$; since $\mathbf{x}_{0}$ is already reintroduced through the
final residual projection, re-feeding it into $\mathcal{B}_{3}$ would be
redundant. The resulting cross-block connections enrich gradient flow and
improve training stability for deep spiking architectures.
 
\subsection{Spike-Domain Decoder and Continuous Reconstruction}

\subsubsection{Spike Pixel-Shuffle Upsampling}

The baseline SSIR uses transposed convolution for spatial recovery.
We instead adopt a learnable sub-pixel upsampling strategy to reconstruct
fine spatial details more effectively and avoid checkerboard artefacts.
Given the bottleneck feature $\mathbf{b}$, we first expand its channel
dimension by a factor of $r^{2}$ (where $r=2$ is the upsampling scale):
\begin{equation}
\mathbf{u}' = \phi\!\left(\text{BN}(\text{Conv}_{3\times3}(\mathbf{b}))\right),
\quad \mathbf{u}'\in\mathbb{R}^{C_{\text{out}}\cdot r^{2}\times\frac{H}{r}\times\frac{W}{r}}.
\end{equation}
The feature is then rearranged by pixel shuffle:
\begin{equation}
\mathbf{u} = \text{PixelShuffle}(\mathbf{u}'),
\quad \mathbf{u}\in\mathbb{R}^{C_{\text{out}}\times H\times W}.
\end{equation}
A lightweight $1\!\times\!1$ refinement layer is further applied:
\begin{equation}
\mathbf{u} \leftarrow \phi\!\left(\text{BN}(\text{Conv}_{1\times1}(\mathbf{u}))\right).
\end{equation}
 
All upsampling coefficients are learned end-to-end, equivalent to a learnable
fractional-stride transposed convolution but without checkerboard artefacts.

\subsubsection{Spike-to-Continuous Fusion Block}

Although the backbone operates in the spiking domain, the final intensity
prediction benefits from a smooth transition from discrete spike activations
to continuous values.
We introduce a fusion block with channel and spatial gating for this purpose.

For an input feature $\mathbf{x}\in\mathbb{R}^{B\times C\times H\times W}$,
the channel gate is
\begin{equation}
\mathbf{g}_{t} = \sigma\!\left(\text{FC}(\text{GAP}(\mathbf{x}))\right)
\in\mathbb{R}^{B\times C},
\end{equation}
and the channel-weighted feature is
\begin{equation}
\mathbf{y}_{t} = \mathbf{x} \odot \mathbf{g}_{t},
\end{equation}
where $\mathbf{g}_{t}$ is broadcast to the spatial dimensions.
The spatial gate uses a large-kernel convolution:
\begin{equation}
\mathbf{g}_{s} = \sigma\!\left(\text{Conv}_{7\times7}(\mathbf{x})\right).
\end{equation}
The spatial branch is then modulated by the complementary factor
$(1-\mathbf{y}_{t})$, which down-weights the content already emphasized by the
channel-gated path so that the two branches attend to complementary information:
\begin{equation}
\mathbf{y}_{s} = \mathbf{x} \odot \mathbf{g}_{s} \odot (1 - \mathbf{y}_{t}).
\end{equation}
The final fused feature sums the two complementary paths:
\begin{equation}
\mathbf{z} = \mathbf{y}_{t} + \mathbf{y}_{s}.
\end{equation}

Through this complementary gating, the channel-weighted path $\mathbf{y}_{t}$
supplies channel-wise global context, while the spatial path $\mathbf{y}_{s}$,
suppressed where $\mathbf{y}_{t}$ is already strong, contributes
spatially-varying structure that the channel gate leaves under-emphasized. The
two paths are thus complementary rather than redundant, providing a smooth
bridge from the discrete spiking backbone to the continuous prediction head.
 
\subsubsection{Prediction Head and Deep Supervision}
 
After upsampling, the decoder feature $\mathbf{u}$ is concatenated with the
refined skip connection from the stem via a $1\!\times\!1$ spiking convolution:
\begin{equation}
\mathbf{f} = \phi\!\left(\text{BN}\!\left(
\text{Conv}_{1\times1}([\mathbf{u},\;\mathcal{R}(\mathbf{s}_{0})])\right)\right),
\end{equation}
where $\mathcal{R}(\cdot)$ denotes the skip-connection refinement layer.
The fused feature is fed into the fusion block to obtain $\mathbf{z}$,
which is then decoded by the lightweight prediction head:
\begin{equation}
\hat{\mathbf{I}} = \sigma\!\left(
\text{Conv}_{1\times1}\!\left(
\phi\!\left(\text{BN}(\text{Conv}_{3\times3}(\mathbf{z}))\right)\right)\right).
\end{equation}
 
During training, an auxiliary prediction head provides deep supervision:
\begin{equation}
\hat{\mathbf{I}}_{\text{aux}} = \sigma\!\left(\text{Conv}_{1\times1}(\mathbf{z})\right).
\end{equation}
The auxiliary output is included directly in the loss without adding inference-time cost.
 
\subsection{Training Objective}
 
For fair comparison with the original SSIR, we adopt the same training
objective.
The loss combines an $\ell_{1}$ reconstruction term and a perceptual term for
both the main and auxiliary predictions.
 
For the main prediction:
\begin{align}
\mathcal{L}_{\text{rec}}^{\text{main}} &=
    \left\|\hat{\mathbf{I}} - \mathbf{I}_{\text{gt}}\right\|_{1},\\
\mathcal{L}_{\text{per}}^{\text{main}} &=
    \left\|\phi_{\text{vgg}}(\hat{\mathbf{I}})
         - \phi_{\text{vgg}}(\mathbf{I}_{\text{gt}})\right\|_{2}^{2},
\end{align}
and for the auxiliary prediction:
\begin{align}
\mathcal{L}_{\text{rec}}^{\text{aux}} &=
    \left\|\hat{\mathbf{I}}_{\text{aux}} - \mathbf{I}_{\text{gt}}\right\|_{1},\\
\mathcal{L}_{\text{per}}^{\text{aux}} &=
    \left\|\phi_{\text{vgg}}(\hat{\mathbf{I}}_{\text{aux}})
         - \phi_{\text{vgg}}(\mathbf{I}_{\text{gt}})\right\|_{2}^{2},
\end{align}
where $\phi_{\text{vgg}}(\cdot)$ denotes feature extraction by a pre-trained
VGG network and $\mathbf{I}_{\text{gt}}$ is the ground-truth intensity frame.
 
The total training loss is
\begin{equation}
\mathcal{L} =
    \mathcal{L}_{\text{rec}}^{\text{main}}
  + \mathcal{L}_{\text{rec}}^{\text{aux}}
  + \lambda\!\left(\mathcal{L}_{\text{per}}^{\text{main}}
                  + \mathcal{L}_{\text{per}}^{\text{aux}}\right),
\end{equation}
where $\lambda=0.2$ is the perceptual loss balancing coefficient.

\section{Experiments}

\subsection{Experimental Settings}

\paragraph{Datasets.}
We evaluate MADSNet on both gray-scale and color spike reconstruction tasks.
For the main gray-scale reconstruction task, we follow SSIR~\cite{zhao2023spike} and conduct
experiments on the simulated SREDS dataset and real-captured spike streams.
SREDS~\cite{zhao2023spike} is generated from REDS~\cite{Nah_2019_CVPR_Workshops_REDS} and contains 524,160 image ground truths and
around $1\times10^{7}$ spike frames.
Following SSIR, we use the first 22 ground-truth frames in the validation
split for evaluation.

To verify the transferability of the proposed design, we further conduct a
color reconstruction experiment on RBSS~\cite{dong2025self} with $\alpha=1.0$ and real-world
Bayer-pattern spike streams from BSS~\cite{dong2024joint}.
This setting allows us to examine whether a spike-based architecture
originally designed for gray-scale reconstruction can be extended to the more
challenging color spike reconstruction scenario.

\paragraph{Implementation Details.}
Unless otherwise specified, we follow the training protocol of SSIR. Each training sample contains 160 spike frames and 8 corresponding ground-truth images. Following SSIR, each of the 8 ground-truth frames is reconstructed from its corresponding local window of spike frames within the 160-frame sample.
Data augmentation includes random horizontal/vertical flipping and
$90^{\circ}/180^{\circ}$ rotations.
We optimize the
network using Adam with an initial learning rate of $4\times10^{-4}$.
The batch size is set to 8.

For the gray-scale task, we crop each scene to $96\times96$ spatially without
overlap, following SSIR.
The learning rate decays every 5 epochs from the 20th to the 70th epoch, and
training runs for 80 epochs in total.

For the color transfer experiment, we crop each scene to $128\times128$
following SSL-CSC~\cite{dong2025self}, and extend the training schedule to
200 epochs while keeping all other hyperparameters unchanged.

\paragraph{Compared Methods.}
For gray-scale reconstruction, we compare MADSNet with representative
conventional and learning-based methods, including TFP~\cite{zhu2019retina},
TFI~\cite{zhu2019retina}, TFSTP~\cite{zheng2021high},
MAHTF~\cite{zhao2021reconstructing},
ETNet~\cite{weng2021event}, HyperE2VID~\cite{ercan2024hypere2vid},
Spk2ImgNet~\cite{zhao2021spk2imgnet}, and SSIR~\cite{zhao2023spike}.
ETNet and HyperE2VID are event-camera reconstruction methods included
as cross-sensor baselines to situate MADSNet within the broader
neuromorphic image reconstruction landscape.
For color reconstruction, we compare against
TFP~\cite{zhu2019retina}, TFI~\cite{zhu2019retina},
TFSTP~\cite{zheng2021high}, 3DRI~\cite{dong20223d},
SJRE~\cite{chen2023self}, SSL-CSC~\cite{dong2025self},
ETNet~\cite{weng2021event}, HyperE2VID~\cite{ercan2024hypere2vid},
Spk2ImgNet~\cite{zhao2021spk2imgnet},
and SSIR~\cite{zhao2023spike} as a direct SNN transfer baseline.

\paragraph{Evaluation Metrics.}
For gray-scale reconstruction on SREDS, we report PSNR, SSIM, and LPIPS~\cite{zhang2018unreasonable},
and additionally report NIQE~\cite{mittal2012making} and BRISQUE~\cite{mittal2012no}.
For color reconstruction on RBSS, we use PSNR, SSIM, and LPIPS.
Lower values of LPIPS, NIQE, and BRISQUE indicate better perceptual quality,
while higher values of PSNR and SSIM are preferred.

\subsection{Comparison on Gray-Scale Spike Reconstruction}

\paragraph{Quantitative Results on SREDS}
Table~\ref{tab:main_gray} compares MADSNet with representative gray-scale
spike reconstruction methods on SREDS. Conventional temporal filtering methods,
including TFP, TFI, and TFSTP, achieve limited reconstruction fidelity, showing
that simple temporal aggregation is insufficient for recovering high-quality
intensity images from sparse spike streams. MAHTF improves the reconstruction
quality by exploiting long-range temporal motion information, but still lags
behind learning-based methods in reference metrics.

Among ANN-based methods, the event-based models ETNet and HyperE2VID adapted for spike streams demonstrate strong capabilities, particularly HyperE2VID achieving 35.456 dB PSNR. Spk2ImgNet similarly achieves a competitive 35.293 dB PSNR. 

Compared with the SNN baseline SSIR, MADSNet improves PSNR from 32.930 dB to
37.232 dB and SSIM from 0.922 to 0.956, while reducing LPIPS from 0.050 to
0.028. This corresponds to a massive +4.30 dB PSNR gain within the SNN paradigm.
Moreover, MADSNet surpasses both the dedicated spike reconstruction ANN (Spk2ImgNet by +1.94 dB) and the modern event-adapted ANN (HyperE2VID by +1.78 dB). These gains demonstrate that a carefully designed spiking architecture can achieve substantially stronger reconstruction fidelity than even advanced conventional ANN baselines. 

For no-reference perceptual metrics, MADSNet obtains competitive NIQE and BRISQUE scores. Although it does not achieve the best value on every
no-reference metric, the overall results show that the proposed architecture
substantially improves reference fidelity while maintaining excellent perceptual quality.

\begin{table}[t]
\centering
\caption{Comparison on the gray-scale spike reconstruction benchmark.
Results of existing methods are taken from SSIR.}
\label{tab:main_gray}
\setlength{\tabcolsep}{2.2pt}
\renewcommand{\arraystretch}{1.03}
\scriptsize
\setlength\tabcolsep{3pt}
\resizebox{\columnwidth}{!}{
\begin{tabular}{l l c c c c c}
\toprule
\multirow{2}{*}{Paradigm} & \multirow{2}{*}{Method}
& \multicolumn{5}{c}{SREDS Dataset} \\
\cmidrule(lr){3-7}
& & PSNR$\uparrow$ & SSIM$\uparrow$ & LPIPS$\downarrow$
& NIQE$\downarrow$ & BRISQUE$\downarrow$ \\
\midrule
\multirow{4}{*}{Non-SNN}
& TFP~\cite{zhu2019retina}
& 28.027 & 0.687 & 0.269 & 7.445 & 40.656 \\
& TFI~\cite{zhu2019retina}
& 24.258 & 0.691 & 0.263 & 5.074 & 40.069 \\
& TFSTP~\cite{zheng2021high}
& 20.845 & 0.640 & 0.271 & 5.163 & 36.948 \\
& MAHTF~\cite{zhao2021reconstructing}
& 31.566 & 0.904 & 0.089 & 2.631 & 15.997 \\
\midrule
\multirow{3}{*}{ANN}
& ETNet~\cite{weng2021event}
& 31.088 & 0.899 & 0.095 & 3.477 & 14.357 \\
& HyperE2VID~\cite{ercan2024hypere2vid}
& 35.456 & 0.932 & 0.069 & 3.338 & 12.636 \\
& Spk2ImgNet~\cite{zhao2021spk2imgnet}
& 35.293 & 0.937 & 0.058 & 3.610 & 20.491 \\
\midrule
\multirow{2}{*}{SNN}
& SSIR~\cite{zhao2023spike}
& 32.930 & 0.922 & 0.050 & 3.111 & 19.694 \\
& MADSNet(Ours)
& \textbf{37.232} & \textbf{0.956} & \textbf{0.028}
& \textbf{2.927} & \textbf{17.550} \\
\bottomrule
\end{tabular}}
\end{table}

\begin{figure*}[t]
    \centering
    \includegraphics[width=1\textwidth]{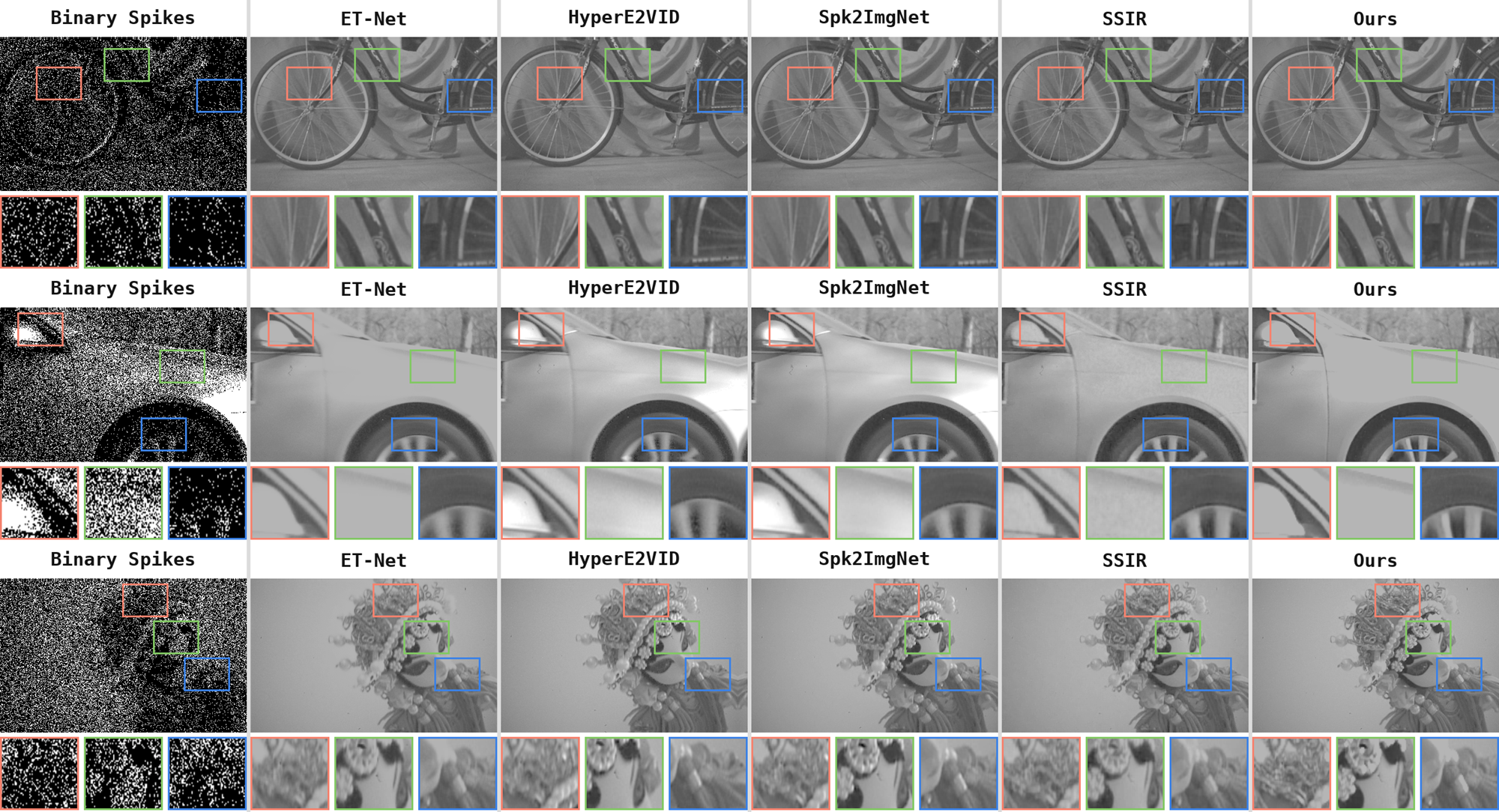}
    \caption{Qualitative comparison on real-captured gray-scale spike streams
    (Binary Spikes, ET-Net, HyperE2VID, Spk2ImgNet, SSIR, Ours).
    Zoom in for more details.
    Gamma transformation with parameter 1.5 is used for visualization.}
    \label{fig:gray_vis}
\end{figure*}

\paragraph{Qualitative Results on Real-Captured Gray-Scale Data.}
Fig.~\ref{fig:gray_vis} presents visual comparisons on real-captured
gray-scale spike streams across three scenes, covering a bicycle, a car,
and a decorative doll.
The event-camera methods ET-Net and HyperE2VID produce visually plausible
reconstructions but tend to introduce blurring or ghosting artifacts in
regions with complex texture, reflecting the inherent limitation of
event cameras in recovering absolute luminance detail.
Spk2ImgNet recovers sharper edges than the event-based methods but still
exhibits residual blur on high-frequency elements.
SSIR reconstructs recognizable structure but loses fine detail, particularly
on thin elements such as bicycle spokes.
MADSNet produces the sharpest and most faithful reconstructions across all
scenes: bicycle spokes are more precisely delineated, high-frequency fan
textures are better preserved with higher local contrast, and fine ornamental
details in the doll scene exhibit fewer boundary artifacts.
These observations are consistent with the quantitative improvements reported
in Table~\ref{tab:main_gray}.

\subsection{Transfer to Color Spike Reconstruction}

Although MADSNet is developed for gray-scale spike reconstruction, its design
is not tied to a specific output modality.
The multi-scale spiking encoder, axial-attentive feature extraction, and
spike-to-continuous decoding are expected to remain useful when the input
changes from gray-scale spike streams to Bayer-pattern color spike streams.
Therefore, we further investigate whether MADSNet can be effectively
transferred to color spike reconstruction, and whether it can serve as a
strong SNN baseline for this more challenging task.

\paragraph{Quantitative Results on RBSS.}
Table~\ref{tab:color_transfer} reports transfer results on the RBSS
color reconstruction benchmark.
Filtering-based methods TFP, TFI, and TFSTP remain below 24 dB PSNR,
confirming their limited capacity to handle the spatial-chromatic coupling
of Bayer-pattern spike streams.
Among event-camera methods, ETNet achieves only 13.10 dB due to the
fundamental modality mismatch between event streams and Bayer spike
streams, while HyperE2VID reaches 27.34 dB, benefiting from its
adaptive per-pixel filtering design.
Spk2ImgNet achieves 29.13 dB, demonstrating the advantage of a
spike-specific ANN architecture.
SSL-CSC attains the highest PSNR (29.95 dB) and SSIM (0.8417) among
all compared methods, benefiting from its motion-guided self-supervised
design tailored for color spike reconstruction.

Directly transferring SSIR to color spike reconstruction yields 24.31 dB
PSNR and 0.6514 SSIM, showing that a spike-based SNN can generalize to the
color domain to a reasonable degree, though with a substantial gap in
fine chromatic detail compared with dedicated methods.
In contrast, MADSNet achieves 29.36 dB PSNR, 0.8434 SSIM, and 0.1464
LPIPS, improving over SSIR by +5.05 dB in PSNR and outperforming all
compared methods in both SSIM and LPIPS.
Notably, MADSNet also surpasses the ANN-based Spk2ImgNet in all three
metrics and closely approaches SSL-CSC in PSNR while outperforming it
in SSIM and LPIPS, despite not being specifically designed for color
reconstruction.
These results demonstrate the strong transferability of the proposed
architecture and confirm that MADSNet serves as a competitive SNN
baseline for color spike reconstruction.


\begin{table}[t]
\centering
\caption{Transferability test on RBSS ($\alpha=1.0$).}
\label{tab:color_transfer}
\setlength\tabcolsep{8pt}
\resizebox{\columnwidth}{!}{
\begin{tabular}{l l ccc}
\toprule
Paradigm & Method & PSNR$\uparrow$ & SSIM$\uparrow$ & LPIPS$\downarrow$ \\
\midrule
\multirow{4}{*}{Non-SNN}
& TFP~\cite{zhu2019retina}   & 23.78 & 0.6207 & 0.3986 \\
& TFI ~\cite{zhu2019retina}  & 23.97 & 0.6294 & 0.3731 \\
& TFSTP~\cite{zheng2021high}   & 23.32 & 0.5978 & 0.3759 \\
& 3DRI~\cite{dong20223d}  & 26.95 & 0.8029 & 0.1946 \\
\midrule
\multirow{5}{*}{ANN}
& SJRE ~\cite{chen2023self}   & 24.84 & 0.5499 & 0.4050 \\
& SSL-CSC~\cite{dong2025self} & 29.95 & 0.8417 & 0.1829 \\
& ETNet~\cite{weng2021event} & 13.10 & 0.5701 & 0.3695 \\
& HyperE2VID~\cite{ercan2024hypere2vid} & 27.34 & 0.7771 & 0.2753 \\
& Spk2ImgNet~\cite{zhao2021spk2imgnet} & 29.13 & 0.8219 & 0.2324 \\
\midrule
\multirow{2}{*}{SNN}
& SSIR~\cite{zhao2023spike}          & 24.31 & 0.6514 & 0.2811 \\
& MADSNet (Ours) & \textbf{29.36} & \textbf{0.8434} & \textbf{0.1464} \\
\bottomrule
\end{tabular}}
\end{table}

\begin{figure*}[t]
    \centering
    \includegraphics[width=1\textwidth]{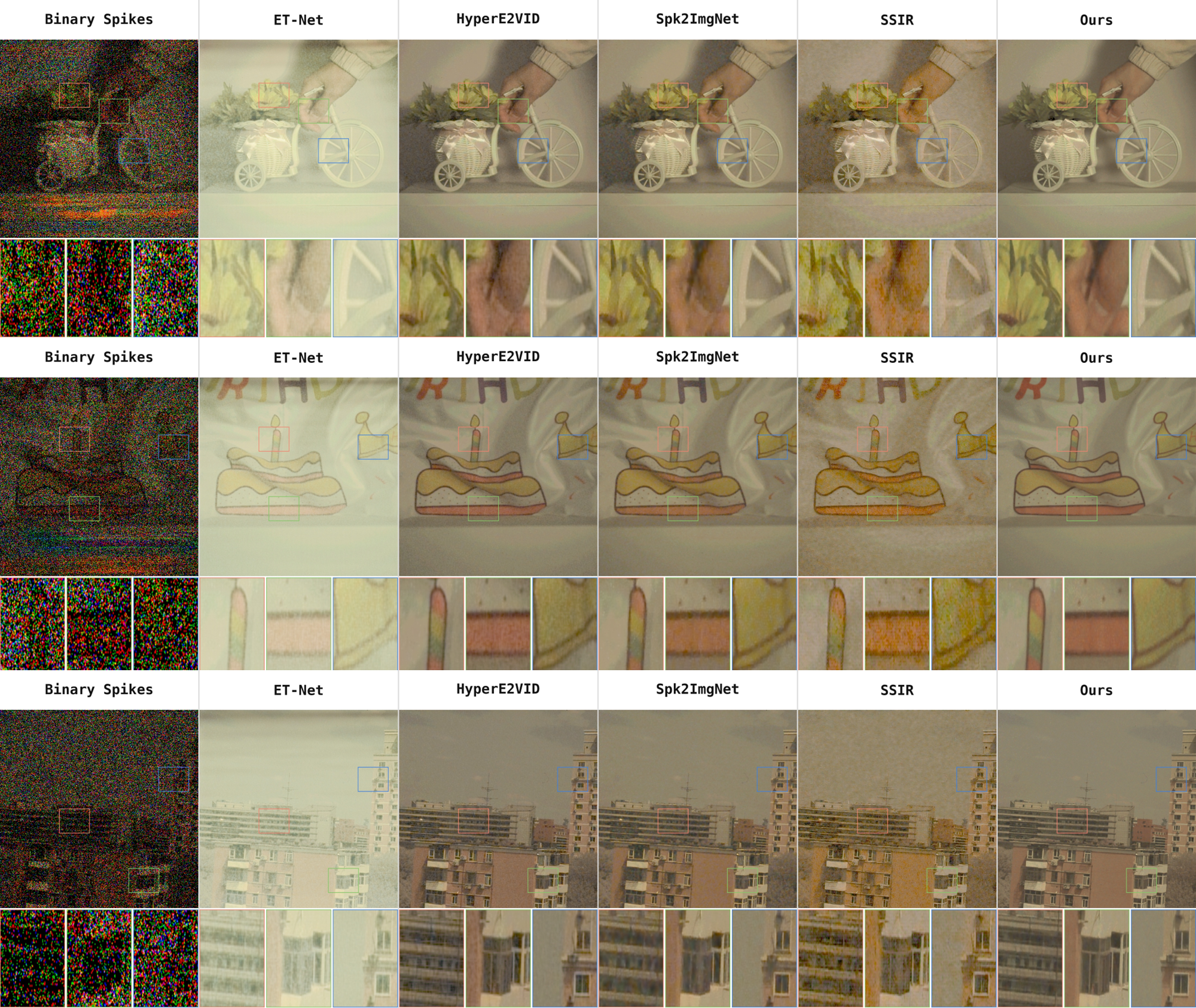}
 \caption{Qualitative comparison on real-captured color spike streams (BSS)
    (Binary Spikes, ET-Net, HyperE2VID, Spk2ImgNet, SSIR, Ours).
    Zoom in for more details.Gamma transformation with parameter 2.2 is used for visualization.}
    \label{fig:color_vis}
\end{figure*}

\paragraph{Qualitative Results on Real-Captured BSS.}
Fig.~\ref{fig:color_vis} presents visual comparisons on real-world BSS
color spike streams across three scenes.
ETNet produces severely overexposed outputs, reflecting the severe modality gap between event cameras and Bayer spike cameras.
HyperE2VID recovers plausible color but introduces noticeable
chromatic noise and loses fine structural boundaries.
Spk2ImgNet reconstructs coherent color images but still exhibits
some color bleeding and texture blurring.
SSIR recovers a coarse color approximation with recognizable scene
structure, though with noticeably reduced chromatic fidelity and
visible mosaic-pattern artifacts attributable to its limited
Bayer-layout modeling.
MADSNet reconstructs well-structured color images with the most
faithful chromatic fidelity and sharpest object boundaries across
all three scenes, consistent with its leading quantitative results
in Table~\ref{tab:color_transfer}.

\paragraph{Analysis of Color Transfer.}
The quantitative and qualitative results together reveal a clear
transferability hierarchy.
SSIR transfers to color reconstruction with moderate success
(24.31 dB), demonstrating that SNN architectures are not
inherently incompatible with the color domain.
However, its single-branch $3\!\times\!3$ spiking stem treats all
stacked spike channels uniformly and provides limited receptive-field
diversity, making it difficult to fully capture chromatic
correlations across the Bayer mosaic.
MADSNet transfers substantially better (+5.05 dB over SSIR)
because its MS3Stem jointly captures inter-channel and multi-scale
spatial patterns, while ASA contributes long-range horizontal and vertical context through its axial modeling.
The spike-to-continuous fusion block further adapts discrete
spike-domain features to the continuous RGB output space.
These components jointly enable MADSNet to recover structured
color images without task-specific architectural redesign,
establishing it as a strong SNN baseline for color spike
reconstruction.

\subsection{Ablation Studies}

\paragraph{Component-Wise Ablation on SREDS}
To validate the individual contribution of each proposed component,
we conduct a systematic ablation study by removing one module at a time
from the full MADSNet and evaluating the resulting model on SREDS.
Table~\ref{tab:ablation} summarizes the results, and we detail the
findings for each component below.

\textbf{MS3Stem.}
Replacing the proposed three-branch multi-scale stem with a single
$3\times3$ spiking convolution (as in SSIR) leads to the most significant
performance drop among all architectural components, reducing PSNR from
37.232 dB to 35.406 dB ($-$1.826 dB) and LPIPS from 0.028 to 0.034.
This large degradation confirms that the rich multi-scale feature embedding
provided by the parallel $1\times1$, $3\times3$, and depthwise-separable
$5\times5$ branches is essential for encoding both temporal inter-channel
correlations and multi-scale spatial patterns from the input spike tensor.
The single-branch baseline is insufficient for capturing the full spatial
and temporal complexity of the stacked spike representation, and the
MS3Stem provides an indispensable foundation for the rest of the network.

\textbf{Axial Spike Attention (ASA).}
Removing ASA from all LightSpikeBlocks results in a PSNR drop from
37.232 dB to 36.601 dB ($-$0.631 dB), with LPIPS increasing from
0.028 to 0.030.
Although the improvement from ASA is more modest than that of MS3Stem,
it is consistent and meaningful. The factorized axial attention provides
the encoder with the ability to model long-range spatial dependencies
along the horizontal and vertical axes at $\mathcal{O}((H+W)\cdot C)$
complexity, which is important for recovering coherent global structures
(e.g., object contours and smooth background regions) in the reconstructed
intensity image.

\textbf{DS-SEWBlock.}
Replacing the proposed depthwise-separable SEW block with the standard
bottleneck SEW block from SSIR reduces PSNR to 36.990 dB ($-$0.242 dB)
and LPIPS to 0.032.
This result confirms that the decoupled spatial and channel mixing
strategy of DS-SEWBlock, together with its inner micro-residual path,
provides benefits beyond simple parameter reduction. The additional
gradient propagation path through the inner micro-residual enhances
feature learning and mitigates potential gradient vanishing in the
depthwise-separable structure.

\textbf{Dense Spike Bottleneck.}
Replacing the dense-connected bottleneck structure with three sequentially
stacked DS-SEWBlocks (removing all inter-block shortcut paths) yields
36.566 dB PSNR ($-$0.666 dB) and 0.032 LPIPS.
The dense connectivity allows $\mathcal{B}_{2}$ to re-access the initial
representation $\mathbf{x}_{0}$ and prevents information forgetting,
while $\mathcal{B}_{3}$ aggregates the adjacent representations
$\mathbf{x}_{1}$ and $\mathbf{x}_{2}$. The additional gradient paths
introduced by these connections also improve training stability in the
deep spiking architecture.
The performance gap confirms that feature reuse in the latent space
is a critical factor for high-fidelity spike camera reconstruction.

\textbf{Spike Pixel-Shuffle Upsampling (SPSU).}
Replacing SPSU with a standard transposed convolution (as in SSIR)
reduces PSNR to 36.581 dB ($-$0.651 dB) and LPIPS to 0.031.
The learnable sub-pixel upsampling avoids the checkerboard artifacts
inherent to transposed convolution and provides a more accurate
reconstruction of fine spatial details, which is particularly important
for recovering high-frequency textures from spike streams.

\textbf{Spike-to-Continuous Fusion Block.}
Removing the fusion block results in a PSNR of 36.085 dB ($-$1.147 dB)
and LPIPS of 0.031, making it the second most impactful component after
MS3Stem. This finding underscores the importance of the smooth transition
from discrete spiking domain features to continuous intensity values.
The joint channel gating and spatially-varying spatial gating of the fusion block enable the network to adaptively weight and recombine
features before the prediction head, providing a critical bridge that
prevents information loss at the spiking-to-continuous interface.

\textbf{Deep Supervision.}
Removing the auxiliary supervision branch yields a modest but consistent
drop (37.232 $\to$ 37.089 dB PSNR, $-$0.143 dB). Although deep supervision
primarily accelerates convergence during training, it also provides
additional gradient paths that serve as implicit regularization,
leading to marginally better generalization at convergence.
Since the auxiliary head is discarded at inference time, this improvement
is obtained at zero additional inference cost.

\textbf{IBRA-LIF vs.\ PLIF Neuron.}
Replacing IBRA-LIF with a standard PLIF neuron causes a catastrophic
performance collapse (37.232 $\to$ 30.230 dB PSNR, $-$7.002 dB;
SSIM: 0.956 $\to$ 0.817; LPIPS: 0.028 $\to$ 0.124).
The nature and significance of this result are discussed in detail
in the following paragraph.

Taken together, these ablation results confirm that every proposed
component makes a meaningful and statistically consistent contribution
to the final reconstruction quality, and that the full MADSNet design
represents an effective and well-motivated integration of all proposed
modules.

\begin{table}[t]
\centering
\caption{Component-wise ablation on SREDS.}
\label{tab:ablation}
\setlength\tabcolsep{4pt}
\resizebox{\columnwidth}{!}{
\begin{tabular}{l ccc}
\toprule
Variant & PSNR$\uparrow$ & SSIM$\uparrow$ & LPIPS$\downarrow$ \\
\midrule
MADSNet (Full)                        & \textbf{37.232} & \textbf{0.956} & \textbf{0.028} \\
\midrule
w/o MS3Stem (single $3\times3$ SConv) & 35.406 & 0.948 & 0.034 \\
w/o ASA                               & 36.601 & 0.955 & 0.030 \\
w/o DS-SEWBlock (standard SEWBlock)   & 36.990 & 0.954 & 0.032 \\
w/o Dense Bottleneck (sequential)     & 36.566 & 0.954 & 0.032 \\
w/o SPSU (transposed conv)            & 36.581 & 0.954 & 0.031 \\
w/o Fusion Block                      & 36.085 & 0.953 & 0.031 \\
w/o Deep Supervision                  & 37.089 & 0.955 & 0.029 \\
PLIF instead of IBRA-LIF              & 30.230 & 0.817 & 0.124 \\
\bottomrule
\end{tabular}}
\end{table}

\paragraph{Neuron-Architecture Interaction}
\begin{table}[t]
\centering
\setlength\tabcolsep{9pt}
\caption{Cross-architecture neuron comparison on SREDS.
IBRA-LIF brings only a marginal gain to SSIR, but is essential for preventing
performance collapse in the deeper MADSNet architecture.}
\label{tab:neuron_cross}
\resizebox{\columnwidth}{!}{
\begin{tabular}{ll ccc}
\toprule
Architecture & Neuron & PSNR$\uparrow$ & SSIM$\uparrow$ & LPIPS$\downarrow$ \\
\midrule
\multirow{2}{*}{SSIR (baseline)}
  & PLIF      & 32.930 & 0.922 & 0.050 \\
  & IBRA-LIF  & 33.168 & 0.937 & 0.030 \\
\midrule
\multirow{2}{*}{MADSNet (Ours)}
  & PLIF      & 30.230 & 0.817 & 0.124 \\
  & IBRA-LIF  & \textbf{37.232} & \textbf{0.956} & \textbf{0.028} \\
\bottomrule
\end{tabular}}
\end{table}
\begin{table}[t]
\centering
\caption{Layer-wise non-zero activation rate (\%) on SREDS validation set.
MADSNet-PLIF exhibits severe activation collapse in deep layers,
while IBRA-LIF maintains healthy activations throughout.}
\label{tab:firing_rate}
\resizebox{\columnwidth}{!}{
\begin{tabular}{lcccc}
\toprule
Layer Group & SSIR & SSIR & MADSNet & MADSNet \\
Neuron & PLIF & IBRA & PLIF & IBRA \\
\midrule
\multicolumn{5}{l}{\textit{SSIR}} \\
Stem (static\_conv)  & 39.66 & 23.53 & -- & -- \\
Encoder (enc1)       & 80.94 & 36.30 & -- & -- \\
Bottleneck (eres1)   & 65.05 & 29.42 & -- & -- \\
Decoder (dec3)       & 83.79 & 13.19 & -- & -- \\
\midrule
\multicolumn{5}{l}{\textit{MADSNet}} \\
MS3Stem              & -- & -- & 6.42  & 35.21 \\
LightSpikeBlock-Stem & -- & -- & 0.19  & 22.31 \\
DownSampling         & -- & -- & 0.02  & 15.43 \\
Dense Bottleneck     & -- & -- & 0.00  & 8.68  \\
SPSU (up)            & -- & -- & 0.74  & 9.40  \\
Fusion Block         & -- & -- & 12.89 & 32.17 \\
Pred Head            & -- & -- & 8.86  & 23.29 \\
\midrule
\textit{Network Avg.} & \textit{67.36} & \textit{25.61} & \textit{4.16} & \textit{20.93} \\
\bottomrule
\end{tabular}}
\end{table}

Although IBRA-LIF is not our architectural contribution, Table~\ref{tab:ablation}
shows that replacing it with PLIF severely degrades MADSNet. We therefore
further analyze whether the gain comes from the neuron alone or from its
interaction with the proposed architecture.

Table~\ref{tab:ablation} shows that replacing IBRA-LIF with
a standard Parametric LIF (PLIF) neuron causes a dramatic
performance drop ($37.232 \to 30.230$ PSNR), which is far
larger than the drops caused by removing any individual
architectural component.
A naive interpretation might attribute this to IBRA-LIF
being the dominant factor; however, this interpretation
is inconsistent with a critical observation: replacing PLIF
with IBRA-LIF in the \emph{original} SSIR yields only a
marginal improvement ($32.930 \to 33.168$ PSNR, $+0.24$ dB),
as shown in Table~\ref{tab:neuron_cross}.
Moreover, MADSNet with PLIF actually \emph{underperforms}
SSIR with PLIF ($30.230$ vs.\ $32.930$), meaning the proposed
architectural improvements are counter-productive when paired
with a neuron that cannot sustain healthy information flow
through a deeper network.

\paragraph{Activation Analysis}

Table~\ref{tab:firing_rate} further reports the layer-wise non-zero activation
rate. MADSNet-PLIF suffers from severe activation collapse in deep stages, with
nearly zero activations in the Dense Bottleneck and only 0.74\% in SPSU.
In contrast, MADSNet-IBRA maintains non-trivial activations across all stages.
This explains why the proposed deeper architecture cannot be effectively
trained with PLIF, whereas IBRA-LIF provides sufficient representational range
to sustain information propagation.

These results indicate that IBRA-LIF is not the sole source of MADSNet's
performance gain. In SSIR, replacing PLIF with IBRA-LIF improves PSNR by only
0.24 dB. With the same IBRA-LIF neuron, MADSNet still outperforms SSIR by
4.06 dB. Therefore, the main gain comes from the proposed architecture, while
IBRA-LIF serves as a necessary activation substrate that enables the deeper
spiking architecture to function properly.

\subsection{Efficiency Analysis}

To evaluate the practical efficiency of MADSNet, we compare it with
Spk2ImgNet (ANN) and SSIR (SNN) in terms of inference latency,
GPU memory footprint, and estimated energy consumption,
measured on an NVIDIA RTX 4090 GPU at $256\times400$ input resolution.
Energy consumption is estimated following the AC/MAC analysis
framework of SSIR~\cite{zhao2023spike} based on per-layer spike firing rates.
The results are summarized in Table~\ref{tab:efficiency}.

The efficiency results highlight two important observations.
First, MADSNet incurs higher latency (20.4 ms vs.\ 3.3 ms),
memory (0.489 GB vs.\ 0.310 GB), and energy (118.4 mJ vs.\ 4.6 mJ)
than the lightweight SSIR baseline. This overhead is the expected
consequence of MADSNet's deeper architecture, which incorporates
multi-scale branching, dense bottleneck connections, and axial
attention modules---all of which contribute to the substantial
performance gains reported in Table~\ref{tab:main_gray}.

The efficiency results show a clear accuracy-efficiency trade-off. Compared
with SSIR, MADSNet has higher latency, memory usage, and estimated energy
consumption, which is expected due to its multi-scale stem, dense bottleneck,
and axial-attentive encoder. In return, MADSNet improves PSNR by +4.30 dB
on SREDS and surpasses the ANN-based Spk2ImgNet in reconstruction fidelity.

More importantly, MADSNet still preserves the efficiency advantage of SNNs over
ANN-based reconstruction. Compared with Spk2ImgNet, MADSNet is approximately
15$\times$ faster, uses 3.4$\times$ less GPU memory, and consumes only about
$1/40$ of the estimated energy. These results indicate that MADSNet achieves a
better balance between reconstruction quality and computational efficiency than
existing ANN and SNN baselines.

\begin{table}[t]
\centering
\setlength\tabcolsep{4pt}
\caption{Efficiency analysis. Spk2ImgNet and SSIR results are taken
from SSIR.}
\label{tab:efficiency}
\resizebox{\columnwidth}{!}{
\begin{tabular}{lcccc}
\toprule
Method & \makecell[c]{Latency $\downarrow$\\(ms)} & \makecell[c]{Memory $\downarrow$\\ (GB)}
       & \makecell[c]{Energy $\downarrow$\\(mJ)} & Norm.\ Energy$\downarrow$ \\
\midrule
Spk2ImgNet     & 305.4 & 1.675 & 4690.0 & 1       \\
SSIR           &   3.3 & 0.310 &    4.6 & 1/1019  \\
MADSNet (Ours) &  20.4 & 0.489 &  118.4 & 1/40    \\
\bottomrule
\end{tabular}}
\end{table}

These advantages stem from the SNN computation paradigm, in which the
binary-range integer spike representation of IBRA-LIF enables the
replacement of multiply-accumulate (MAC) operations with efficient
accumulate-only (AC) operations during inference, substantially
reducing the effective arithmetic cost per layer.
Furthermore, the depthwise-separable structure of DS-SEWBlock
contributes to a reduced parameter count and lower activation volume,
partially mitigating the additional overhead introduced by the deeper
architecture compared to SSIR.

\section{Conclusion}
In this paper, we proposed MADSNet, a multi-scale axial-attentive dense spiking network for spike camera image reconstruction. MADSNet is designed to better exploit the complementary advantages of spike cameras and SNNs: spike cameras provide high-temporal-resolution binary visual streams, while SNNs enable efficient spike-domain computation. To improve reconstruction fidelity beyond existing shallow SNN designs, MADSNet enriches the spike-domain representation along the full encoder--decoder pipeline, from multi-scale early embedding, through axial-attentive encoding and dense latent aggregation, to spike-to-continuous decoding. 

Extensive experiments on gray-scale and color spike reconstruction demonstrate the effectiveness and generality of the proposed architecture. On SREDS, MADSNet significantly outperforms the SNN baseline SSIR and surpasses the ANN-based Spk2ImgNet in reconstruction accuracy, while maintaining much lower latency and estimated energy consumption than Spk2ImgNet. Color reconstruction experiments show that MADSNet transfers better than the direct SNN baseline. These results suggest that carefully designed SNN architectures can provide a favorable accuracy--efficiency trade-off for spike camera reconstruction. Nevertheless, MADSNet's deeper architecture incurs higher latency and energy than lightweight shallow SNNs such as SSIR; reducing this overhead, e.g., through neuromorphic-hardware deployment or further architectural slimming, is a promising direction for future work.

\bibliographystyle{IEEEtran}
\bibliography{refs}

@inproceedings{zhao2024recognizing,
  title={Recognizing ultra-high-speed moving objects with bio-inspired spike camera},
  author={Zhao, Junwei and Zhang, Shiliang and Yu, Zhaofei and Huang, Tiejun},
  booktitle={Proceedings of the AAAI Conference on Artificial Intelligence},
  volume={38},
  number={7},
  pages={7478--7486},
  year={2024}
}

@article{zheng2022spike,
  title={Spike-based motion estimation for object tracking through bio-inspired unsupervised learning},
  author={Zheng, Yajing and Yu, Zhaofei and Wang, Song and Huang, Tiejun},
  journal={IEEE Transactions on Image Processing},
  volume={32},
  pages={335--349},
  year={2022},
  publisher={IEEE}
}

@article{gallego2020event,
  title={Event-based vision: A survey},
  author={Gallego, Guillermo and Delbr{\"u}ck, Tobi and Orchard, Garrick and Bartolozzi, Chiara and Taba, Brian and Censi, Andrea and Leutenegger, Stefan and Davison, Andrew J and Conradt, J{\"o}rg and Daniilidis, Kostas and others},
  journal={IEEE transactions on pattern analysis and machine intelligence},
  volume={44},
  number={1},
  pages={154--180},
  year={2020},
  publisher={IEEE}
}

@article{zheng2026snntracker,
  author={Zheng, Yajing and Li, Chengen and Zhang, Jiyuan and Yu, Zhaofei and Huang, Tiejun},
  journal={IEEE Transactions on Pattern Analysis and Machine Intelligence}, 
  title={SNNTracker: Online High-Speed Multi-Object Tracking With Spike Camera}, 
  year={2026},
  volume={48},
  number={1},
  pages={624-638},
doi={10.1109/TPAMI.2025.3610696}}

@inproceedings{chen2025spikeclip,
  title={Rethinking high-speed image reconstruction framework with spike camera},
  author={Chen, Kang and Zheng, Yajing and Huang, Tiejun and Yu, Zhaofei},
  booktitle={Proceedings of the AAAI Conference on Artificial Intelligence},
  volume={39},
  number={2},
  pages={2097--2104},
  year={2025}
}

@article{huang20231000,
  title={1000$\times$ faster camera and machine vision with ordinary devices},
  author={Huang, Tiejun and Zheng, Yajing and Yu, Zhaofei and Chen, Rui and Li, Yuan and Xiong, Ruiqin and Ma, Lei and Zhao, Junwei and Dong, Siwei and Zhu, Lin and others},
  journal={Engineering},
  volume={25},
  pages={110--119},
  year={2023},
  publisher={Elsevier}
}

@article{zheng2026spikecv,
  title={{SpikeCV}: open a continuous computer vision era},
  author={Zheng, Yajing and Zhang, Jiyuan and Zhao, Rui and Ding, Jianhao and Chen, Shiyan and Wu, Weijian and Xiong, Ruiqin and Yu, Zhaofei and Huang, Tiejun},
  journal={Science China: Information Sciences},
  volume={69},
  number={3},
  pages={132106},
  year={2026},
  publisher={Springer}
}

@InProceedings{zhu2019retina,
  author    = {Zhu, Lin and Dong, Siwei and Huang, Tiejun and Tian, Yonghong},
  booktitle = {IEEE International Conference on Multimedia and Expo},
  title     = {A retina-inspired sampling method for visual texture reconstruction},
  year      = {2019},
}

@article{zheng2023capture,
  title={Capture the moment: High-speed imaging with spiking cameras through short-term plasticity},
  author={Zheng, Yajing and Zheng, Lingxiao and Yu, Zhaofei and Huang, Tiejun and Wang, Song},
  journal={IEEE Transactions on Pattern Analysis and Machine Intelligence},
  volume={45},
  number={7},
  pages={8127--8142},
  year={2023},
  publisher={IEEE}
}

@inproceedings{zhao2021spk2imgnet,
  title={Spk2imgnet: Learning to reconstruct dynamic scene from continuous spike stream},
  author={Zhao, Jing and Xiong, Ruiqin and Liu, Hangfan and Zhang, Jian and Huang, Tiejun},
  booktitle={IEEE/CVF Conference on Computer Vision and Pattern Recognition},
  pages={11996--12005},
  year={2021}
}

@inproceedings{chen2022self,
  title={Self-Supervised Mutual Learning for Dynamic Scene Reconstruction of Spiking Camera.},
  author={Chen, Shiyan and Duan, Chaoteng and Yu, Zhaofei and Xiong, Ruiqin and Huang, Tiejun},
  booktitle={International Joint Conference on Artificial Intelligence},
  pages={2859--2866},
  year={2022}
}

@article{zhao2023spike,
  title={Spike camera image reconstruction using deep spiking neural networks},
  author={Zhao, Rui and Xiong, Ruiqin and Zhang, Jian and Yu, Zhaofei and Zhu, Shuyuan and Ma, Lei and Huang, Tiejun},
  journal={IEEE Transactions on Circuits and Systems for Video Technology},
  volume={34},
  number={6},
  pages={5207--5212},
  year={2023},
  publisher={IEEE}
}

@inproceedings{zhao2024boosting,
  title={Boosting spike camera image reconstruction from a perspective of dealing with spike fluctuations},
  author={Zhao, Rui and Xiong, Ruiqin and Zhao, Jing and Zhang, Jian and Fan, Xiaopeng and Yu, Zhaofei and Huang, Tiejun},
  booktitle={IEEE/CVF Conference on Computer Vision and Pattern Recognition},
  pages={24955--24965},
  year={2024}
}

@inproceedings{zheng2021high,
  title={High-speed image reconstruction through short-term plasticity for spiking cameras},
  author={Zheng, Yajing and Zheng, Lingxiao and Yu, Zhaofei and Shi, Boxin and Tian, Yonghong and Huang, Tiejun},
  booktitle={Proceedings of the IEEE/CVF Conference on Computer Vision and Pattern Recognition},
  pages={6358--6367},
  year={2021}
}

@article{zhao2021reconstructing,
  title={Reconstructing clear image for high-speed motion scene with a retina-inspired spike camera},
  author={Zhao, Jing and Xiong, Ruiqin and Xie, Jiyu and Shi, Boxin and Yu, Zhaofei and Gao, Wen and Huang, Tiejun},
  journal={IEEE Transactions on Computational Imaging},
  volume={8},
  pages={12--27},
  year={2021},
  publisher={IEEE}
}

@inproceedings{dong20223d,
  title={3D residual interpolation for spike camera demosaicing},
  author={Dong, Yanchen and Zhao, Jing and Xiong, Ruiqin and Huang, Tiejun},
  booktitle={2022 IEEE International Conference on Image Processing (ICIP)},
  pages={1461--1465},
  year={2022},
  organization={IEEE}
}

@inproceedings{chen2023self,
  title={Self-supervised joint dynamic scene reconstruction and optical flow estimation for spiking camera},
  author={Chen, Shiyan and Yu, Zhaofei and Huang, Tiejun},
  booktitle={Proceedings of the AAAI Conference on Artificial Intelligence},
  volume={37},
  number={1},
  pages={350--358},
  year={2023}
}

@inproceedings{dong2025self,
  title={Self-supervised learning for color spike camera reconstruction},
  author={Dong, Yanchen and Xiong, Ruiqin and Fan, Xiaopeng and Yu, Zhaofei and Tian, Yonghong and Huang, Tiejun},
  booktitle={Proceedings of the Computer Vision and Pattern Recognition Conference},
  pages={6231--6240},
  year={2025}
}

@article{maass1997networks,
  title={Networks of spiking neurons: the third generation of neural network models},
  author={Maass, Wolfgang},
  journal={Neural networks},
  volume={10},
  number={9},
  pages={1659--1671},
  year={1997},
  publisher={Elsevier}
}

@article{wu2018spatio,
  title={Spatio-temporal backpropagation for training high-performance spiking neural networks},
  author={Wu, Yujie and Deng, Lei and Li, Guoqi and Zhu, Jun and Shi, Luping},
  journal={Frontiers in Neuroscience},
  volume={12},
  pages={331},
  year={2018},
  publisher={Frontiers Media SA}
}

@inproceedings{fang2021incorporating,
  title={Incorporating learnable membrane time constant to enhance learning of spiking neural networks},
  author={Fang, Wei and Yu, Zhaofei and Chen, Yanqi and Masquelier, Timoth{\'e}e and Huang, Tiejun and Tian, Yonghong},
  booktitle={Proceedings of the IEEE/CVF international conference on computer vision},
  pages={2661--2671},
  year={2021}
}

@article{li2021differentiable,
  title={Differentiable spike: Rethinking gradient-descent for training spiking neural networks},
  author={Li, Yuhang and Guo, Yufei and Zhang, Shanghang and Deng, Shikuang and Hai, Yongqing and Gu, Shi},
  journal={Advances in neural information processing systems},
  volume={34},
  pages={23426--23439},
  year={2021}
}

@inproceedings{luo2024integer,
  title={Integer-valued training and spike-driven inference spiking neural network for high-performance and energy-efficient object detection},
  author={Luo, Xinhao and Yao, Man and Chou, Yuhong and Xu, Bo and Li, Guoqi},
  booktitle={European Conference on Computer Vision},
  pages={253--272},
  year={2024},
  organization={Springer}
}

@article{ye2025integer,
  title={Integer Binary-Range Alignment Neuron for Spiking Neural Networks},
  author={Ye, Binghao and Li, Wenjuan and Wang, Dong and Yao, Man and Li, Bing and Hu, Weiming and Liang, Dong and Shang, Kun},
  journal={arXiv preprint arXiv:2506.05679},
  year={2025}
}

@inproceedings{Nah_2019_CVPR_Workshops_REDS,
  title={Ntire 2019 challenge on video deblurring and super-resolution: Dataset and study},
  author={Nah, Seungjun and Baik, Sungyong and Hong, Seokil and Moon, Gyeongsik and Son, Sanghyun and Timofte, Radu and Mu Lee, Kyoung},
  booktitle={Proceedings of the IEEE/CVF conference on computer vision and pattern recognition workshops},
  pages={0--0},
  year={2019}
}

@inproceedings{dong2024joint,
  title={Joint demosaicing and denoising for spike camera},
  author={Dong, Yanchen and Xiong, Ruiqin and Zhao, Jing and Zhang, Jian and Fan, Xiaopeng and Zhu, Shuyuan and Huang, Tiejun},
  booktitle={Proceedings of the AAAI Conference on Artificial Intelligence},
  volume={38},
  number={2},
  pages={1582--1590},
  year={2024}
}

@inproceedings{zhang2018unreasonable,
  title={The unreasonable effectiveness of deep features as a perceptual metric},
  author={Zhang, Richard and Isola, Phillip and Efros, Alexei A and Shechtman, Eli and Wang, Oliver},
  booktitle={Proceedings of the IEEE conference on computer vision and pattern recognition},
  pages={586--595},
  year={2018}
}

@article{mittal2012making,
  title={Making a “completely blind” image quality analyzer},
  author={Mittal, Anish and Soundararajan, Rajiv and Bovik, Alan C},
  journal={IEEE Signal Processing Letters},
  volume={20},
  number={3},
  pages={209--212},
  year={2012},
}

@article{mittal2012no,
  title={No-reference image quality assessment in the spatial domain},
  author={Mittal, Anish and Moorthy, Anush Krishna and Bovik, Alan Conrad},
  journal={IEEE Transactions on image processing},
  volume={21},
  number={12},
  pages={4695--4708},
  year={2012},
  publisher={IEEE}
}

@inproceedings{weng2021event,
  title={Event-based video reconstruction using transformer},
  author={Weng, Wenming and Zhang, Yueyi and Xiong, Zhiwei},
  booktitle={Proceedings of the IEEE/CVF International Conference on Computer Vision},
  pages={2563--2572},
  year={2021}
}

@article{ercan2024hypere2vid,
  title={HyperE2VID: Improving event-based video reconstruction via hypernetworks},
  author={Ercan, Burak and Eker, Onur and Saglam, Canberk and Erdem, Aykut and Erdem, Erkut},
  journal={IEEE Transactions on Image Processing},
  volume={33},
  pages={1826--1837},
  year={2024},
  publisher={IEEE}
}

@inproceedings{rebecq2019events,
  title={Events-to-video: Bringing modern computer vision to event cameras},
  author={Rebecq, Henri and Ranftl, Ren{\'e} and Koltun, Vladlen and Scaramuzza, Davide},
  booktitle={Proceedings of the IEEE/CVF conference on computer vision and pattern recognition},
  pages={3857--3866},
  year={2019}
}

@inproceedings{scheerlinck2020fast,
  title={Fast image reconstruction with an event camera},
  author={Scheerlinck, Cedric and Rebecq, Henri and Gehrig, Daniel and Barnes, Nick and Mahony, Robert and Scaramuzza, Davide},
  booktitle={Proceedings of the IEEE/CVF winter conference on applications of computer vision},
  pages={156--163},
  year={2020}
}

@inproceedings{stoffregen2020reducing,
  title={Reducing the sim-to-real gap for event cameras},
  author={Stoffregen, Timo and Scheerlinck, Cedric and Scaramuzza, Davide and Drummond, Tom and Barnes, Nick and Kleeman, Lindsay and Mahony, Robert},
  booktitle={European Conference on Computer Vision},
  pages={534--549},
  year={2020},
  organization={Springer}
}

@inproceedings{zhang2024spike,
  title={Spike-guided motion deblurring with unknown modal spatiotemporal alignment},
  author={Zhang, Jiyuan and Chen, Shiyan and Zheng, Yajing and Yu, Zhaofei and Huang, Tiejun},
  booktitle=CVPR,
  pages={25047--25057},
  year={2024}
}

@article{zhao2025spike,
  title={Spike Camera Optical Flow Estimation Based on Continuous Spike Streams},
  author={Zhao, Rui and Xiong, Ruiqin and Wang, Dongkai and Xuan, Shiyu and Zhang, Jian and Fan, Xiaopeng and Huang, Tiejun},
  journal=TPAMI,
  year={2026},
  volume={48},
  number={4},
  pages={4756-4770},
}

@inproceedings{zhao2024optical,
  title={Optical flow for spike camera with hierarchical spatial-temporal spike fusion},
  author={Zhao, Rui and Xiong, Ruiqin and Zhang, Jian and Zhang, Xinfeng and Yu, Zhaofei and Huang, Tiejun},
  booktitle=AAAI,
  pages={7496--7504},
  year={2024}
}

@inproceedings{zhao2022learning,
  title={Learning optical flow from continuous spike streams},
  author={Zhao, Rui and Xiong, Ruiqin and Zhao, Jing and Yu, Zhaofei and Fan, Xiaopeng and Huang, Tiejun},
  booktitle=NeurIPS,
  pages={7905--7920},
  year={2022}
}

@inproceedings{hu2022optical,
  title={Optical flow estimation for spiking camera},
  author={Hu, Liwen and Zhao, Rui and Ding, Ziluo and Ma, Lei and Shi, Boxin and Xiong, Ruiqin and Huang, Tiejun},
  booktitle={Proceedings of the IEEE/CVF conference on computer vision and pattern recognition},
  pages={17844--17853},
  year={2022}
}

@article{zhu2022fpga,
  title={An FPGA accelerator for high-speed moving objects detection and tracking with a spike camera},
  author={Zhu, Yaoyu and Zhang, Yu and Xie, Xiaodong and Huang, Tiejun},
  journal={Neural Computation},
  volume={34},
  number={8},
  pages={1812--1839},
  year={2022},
}

@article{dong2025color,
  title={Color Spike Camera Reconstruction via Long Short-Term Temporal Aggregation of Spike Signals},
  author={Dong, Yanchen and Xiong, Ruiqin and Zhao, Jing and Fan, Xiaopeng and Zhang, Xinfeng and Huang, Tiejun},
  journal={IEEE Transactions on Image Processing},
  year={2025},
  publisher={IEEE}
}

@article{dong2024learning,
  title={Learning a deep demosaicing network for spike camera with color filter array},
  author={Dong, Yanchen and Xiong, Ruiqin and Zhao, Jing and Zhang, Jian and Fan, Xiaopeng and Zhu, Shuyuan and Huang, Tiejun},
  journal={IEEE Transactions on Image Processing},
  volume={33},
  pages={3634--3647},
  year={2024},
  publisher={IEEE}
}

@article{feng2025joint,
  title={A Joint Visual Compression and Perception Framework for Neuromorphic Spiking Camera},
  author={Feng, Kexiang and Jia, Chuanmin and Ma, Siwei and Gao, Wen},
  journal={IEEE Transactions on Image Processing},
  year={2025},
  publisher={IEEE}
}

\vspace{-4\baselineskip}
\begin{IEEEbiography}[{\includegraphics[width=1in,height=1.25in,clip,keepaspectratio]{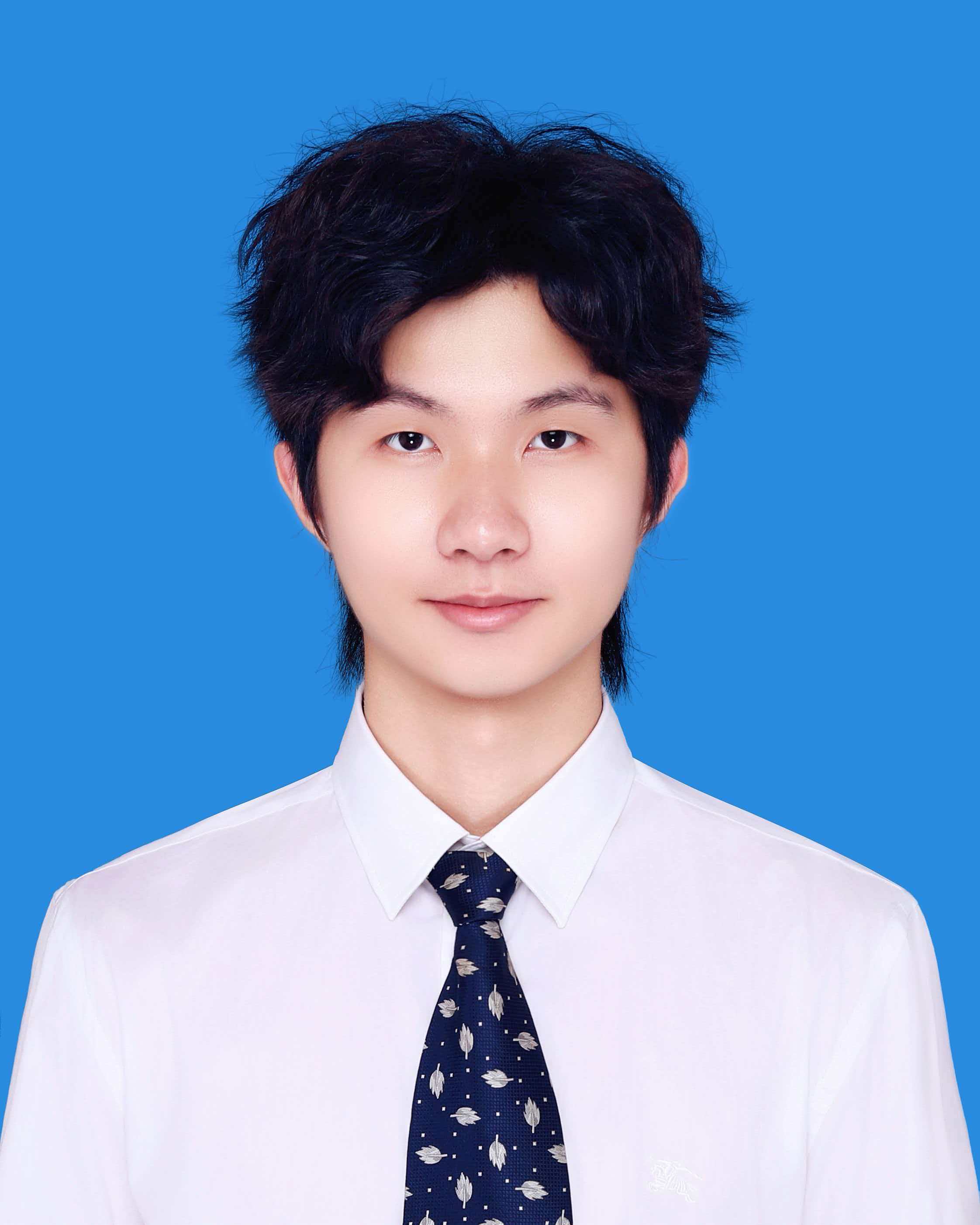}}]{Sirui Li}
 received the B.S. degree from the School of Mathematical Sciences, Dalian University of Technology, Dalian, China, in 2023, and the M.S. degree from the School of Data Science, The Chinese University of Hong Kong, Shenzhen, China, in 2025. He is currently pursuing the Ph.D. degree with the School of Data Science, The Chinese University of Hong Kong, Shenzhen, China. His research interests include spiking neural networks and their applications in image, EEG, and speech signal processing.
\end{IEEEbiography}
\vspace{-3\baselineskip}
\begin{IEEEbiography}[{\includegraphics[width=1in,height=1.25in,clip,keepaspectratio]{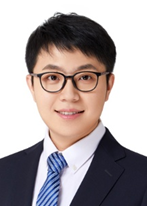}}]{Rui Zhao}
 (Member, IEEE) received the B.E. degree in Communication Engineering from Tianjin University, Tianjin, China, in 2020, and the Ph.D. degree in Computer Science from Peking University, Beijing, China, in 2025. He is currently a postdoctoral research fellow with Nanyang Technological University, Singapore. His research interests include computational photography, computer vision, and image processing, particularly for topics related to motion estimation and neuromorphic cameras.
\end{IEEEbiography}
\begin{IEEEbiography}[{\includegraphics[width=1in,height=1.25in,clip,keepaspectratio]{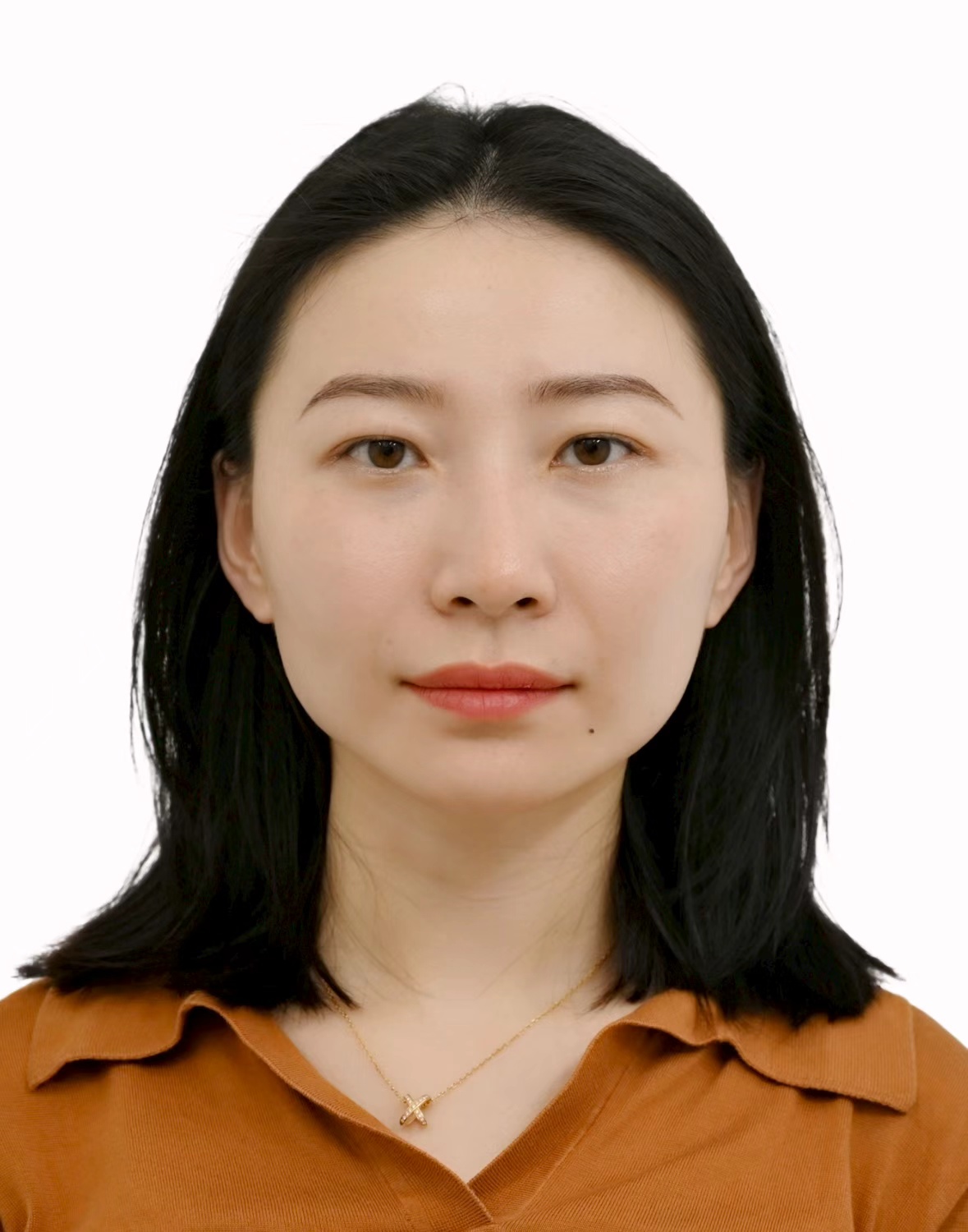}}]{Siqi Cai}
 (Senior Member, IEEE) received the Ph.D. degree in intelligent engineering from South China University of Technology, Guangzhou, China, in 2020. She is currently a Professor with Harbin Institute of Technology, Shenzhen, China. From 2020 to 2024, she was a Research Fellow with the Department of Electrical and Computer Engineering, National University of Singapore, Singapore. Her research interests include brain--computer interfaces and multimodal biosignal processing. She served as Workshop Chair of ICASSP 2022 and Entrepreneurship Forum Chair of ICASSP 2024 and ICASSP 2026, and is an Associate Editor of the International Journal of Social Robotics. She has been a Council Member of the Chinese and Oriental Languages Information Processing Society since 2022.
\end{IEEEbiography}
\begin{IEEEbiography}[{\includegraphics[width=1in,height=1.25in,clip,keepaspectratio]{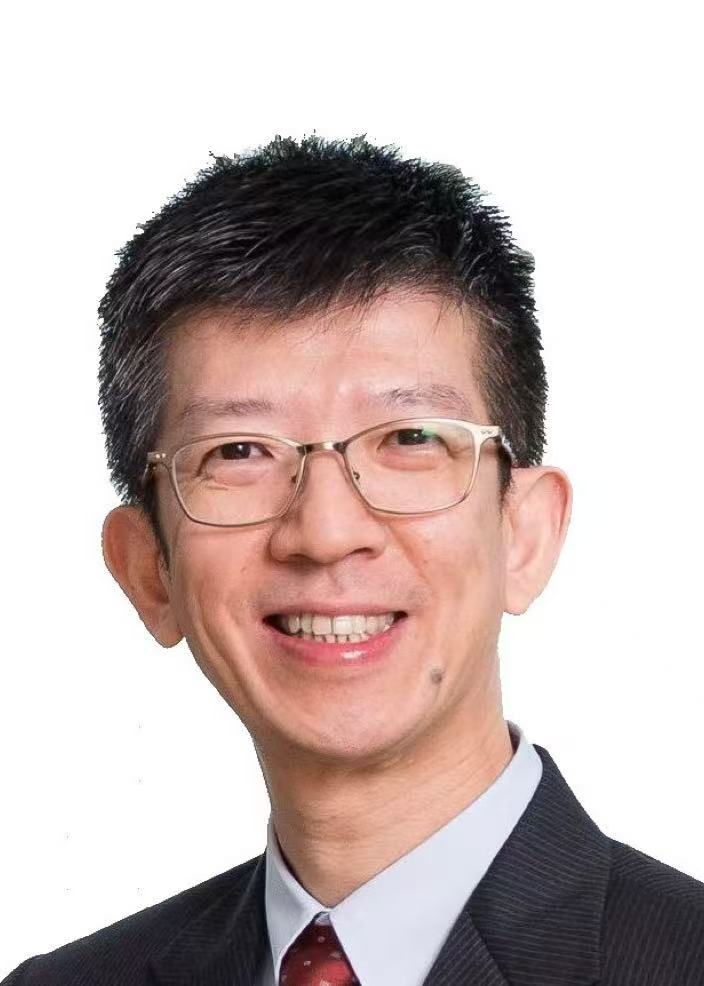}}]{Haizhou Li}
(Fellow, IEEE) received the B.Sc., M.Sc., and Ph.D. degrees in electrical and electronic
engineering from South China University of Technology, Guangzhou, China in 1984, 1987, and 1990 respectively. He is currently a Presidential Chair Professor with the School of Data Science, The Chinese University of Hong Kong, Shenzhen, China. He is also an Adjunct Professor with the Department of Electrical and Computer Engineering, National University of Singapore (NUS). His research interests include automatic speech recognition, speaker and language recognition, natural language processing. Dr Li was the Editor-in-Chief of IEEE/ACM Transactions on Audio, Speech and Language Processing from 2015 to 2018, a Member of the Editorial Board of Computer Speech and Language since 2012, an elected Member of IEEE Speech and Language Processing Technical Committee during 2013–2015, the President of the International Speech Communication Association during 2015–2017, the President of Asia Pacific Signal and Information Processing Association during 2015--2016, the President of Asian Federation of Natural Language Processing during 2017--2018, and Vice President of IEEE Signal Processing Society during 2024--2026. He was the General Chair of ACL 2012, INTERSPEECH 2014, ASRU 2019 and ICASSP 2022. Dr. Li is a Fellow of the ISCA, and a Fellow of the Academy of Engineering Singapore.
\end{IEEEbiography}
\begin{IEEEbiography}[{\includegraphics[width=1in,height=1.25in,clip,keepaspectratio]{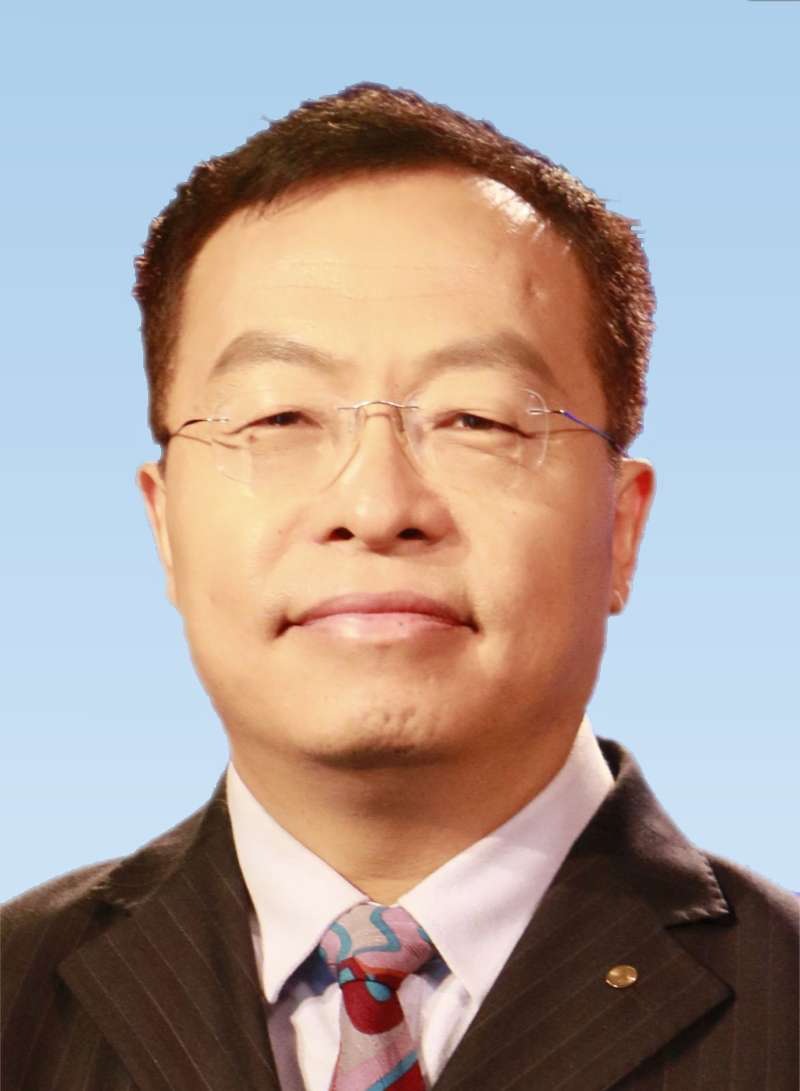}}]{Tiejun Huang}
(Senior Member, IEEE) is currently a Professor with the School of Computer Science, Peking University, Beijing, China, and the Dean of Beijing Academy of Artificial Intelligence (BAAI). He is a recipient of the National Science Fund for Distinguished Young Scholars, Changjiang Distinguished Professorship, and National Talent Program for Technological Innovation Leadership. He received the Ph.D. degree in pattern recognition and intelligent systems from Huazhong (Central China) University of Science and Technology, Wuhan, China, in 1998, and the master’s and bachelor’s degree in computer science from the Wuhan University of Technology, Wuhan, in 1995 and 1992, respectively. His research area includes video coding, image understanding, digital right management, and digital library. He has authored or co-authored over 100 peer-reviewed papers and three books.
\end{IEEEbiography}
\begin{IEEEbiography}[{\includegraphics[width=1in,height=1.25in,clip,keepaspectratio]{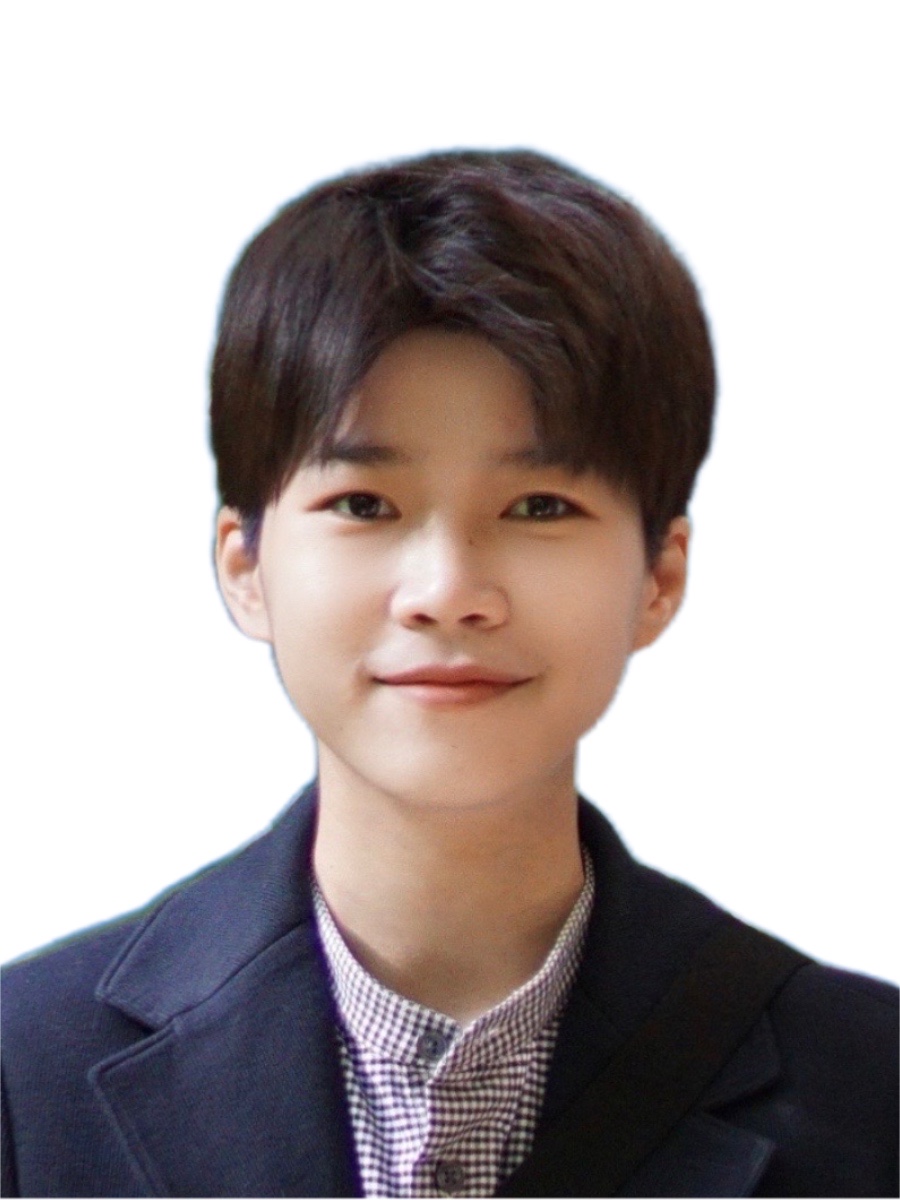}}]{Yajing Zheng}
(Member, IEEE) received the B.S. degree from Sichuan University, Chengdu, China, in 2017, and the Ph.D. degree from the School of Computer Science, Peking University, Beijing, China, in 2022. She is currently a Boya Postdoctoral Fellow with the School of Computer Science, Peking University, and an Honorary Research Fellow with UCL Imaging Neuroscience, University College London, London, U.K. Her research focuses on neuromorphic vision and brain-inspired computing, with particular interests in spike cameras, event-based vision, spiking neural networks, computational neuroscience, and low-latency intelligent perception systems. She serves as the Deputy Secretary-General of the Brain-Inspired Vision Committee of the China Society of Image and Graphics.
\end{IEEEbiography}
\end{document}